\documentclass[twocolumn]{article}
\usepackage{graphicx}
\usepackage{subcaption}
\usepackage{multirow, makecell}
\usepackage{authblk}
\usepackage[square,numbers]{natbib}
\usepackage{hyperref}
\hypersetup{pdfborder={0 0 0}}
\usepackage[acronym]{glossaries}
\usepackage{enumitem}
\makeglossaries
\usepackage{fancyhdr}
\date{}
\newcommand{\Q}[1]{`#1'}
\newcommand{\fig}[1]{\mbox{Figure \ref{#1}}}
\newcommand{\tab}[1]{\mbox{Table \ref{#1}}}
\newcommand{\secr}[1]{\mbox{\S\ref{#1}}}
\newcommand{\eq}[1]{\mbox{Equation \ref{#1}}}
\newcommand{\F}{\mbox{F\textsubscript{1}-score}}
\newcommand{\E}{\mbox{Edit-score}}

\setacronymstyle{long-short}
\newacronym{EV}{EndoVis}{Endoscopic Vision}
\newacronym{eTSA}{eTSA}{endoscopic transsphenoidal approach}
\newacronym{fps}{FPS}{Frames Per Second}
\newacronym{ML}{ML}{Machine Learning}
\newacronym{MICCAI}{MICCAI}{Medical Image Computing and Computer Assisted Interventions}
\newacronym{PV}{PitVis}{Pituitary Vision}
\newacronym{CRUK}{CRUK}{Cancer Research UK}
\newacronym{DSIT}{DSIT}{Department of Science, Innovation and Technology}
\newacronym{EPSRC}{EPSRC}{Engineering and Physical Sciences Research Council}
\newacronym{NHNN}{NHNN}{National Hospital for Neurology and Neurosurgery}
\newacronym{NIHR}{NIHR}{National Institute for Health and Care Research}
\newacronym{UCL}{UCL}{University College London}
\newacronym{WEISS}{WEISS}{Wellcome/EPSRC Centre for Interventional and Surgical Sciences}
\newacronym{UK}{UK}{United Kingdom}
\newacronym{CNN}{CNN}{Convolution Neural Network}
\newacronym{GRU}{GRU}{Gated Recurrent Unit}
\newacronym{HMM}{HMM}{Hidden Markov Model}
\newacronym{LSTM}{LSTM}{Long Short Term Memory Network}
\newacronym{MHSA}{MHSA}{Multi-Head Self-Attention}
\newacronym{MMHA}{MMHA}{Masked Multi-Head Attention}
\newacronym{RNN}{RNN}{Recurrent Neural Network}
\newacronym{TCN}{TCN}{Temporal Convolution Neural Network}
\newacronym{TSF}{TSF}{Temporal Smoothing Function}
\newacronym{SSM}{SSM}{Sufficient Statistics Model}
\newacronym{S-E}{S-E}{Spatial Encoder}
\newacronym{ST-TF}{ST-TF}{Spatio-Temporal Transformer}
\newacronym{ST-E}{ST-E}{Spatio-Temporal Encoder}
\newacronym{ST-D}{ST-D}{Spatio-Temporal Decoder}
\newacronym{S-TF}{S-TF}{Spatial Transformer}
\newacronym{T-TF}{T-TF}{Temporal Transformer}
\newacronym{Adam}{Adam}{Adaptive Moment Estimation}
\newacronym{BCE}{BCE}{Binary Cross-Entropy Loss Function}
\newacronym{CE}{CE}{Cross-Entropy Loss Function}
\newacronym{conv}{conv}{convolution layers}
\newacronym{CLAHE}{CLAHE}{Contrast Limited Adaptive Histogram Equalization}
\newacronym{ETE}{ETE}{End to End Temporal Training}
\newacronym{GeLU}{GeLU}{Gaussian error Linear Unit}
\newacronym{HSV}{HSV}{Hue Saturation Value}
\newacronym{IRB}{IRB}{Institutional Review Board}
\newacronym{mAP}{mAP}{mean Average Precision}
\newacronym{ReLU}{ReLU}{Rectified Linear Unit}
\newacronym{RBG}{RBG}{Red Blue Green}
\newacronym{Sep}{Sep}{Seperate Temporal Training}
\newacronym{std}{std}{Standard Deviation}
\newacronym{SGD}{SGD}{Stochastic Gradient Descent}
\newacronym{TS}{TS}{Temporal Smoothing Loss Function}
\newacronym{Val}{Val}{Validation Dataset}

\title{
\rule{\linewidth}{3pt}
\textbf{PitVis-2023 Challenge:} 
\\Workflow Recognition in videos of Endoscopic Pituitary Surgery
\rule{\linewidth}{3pt}
}
\author[1,+]{\small Adrito Das}
\author[1,2]{Danyal Z. Khan}
\author[1]{Dimitrios Psychogyios}
\author[1]{Yitong Zhang}
\author[1,2]{John G. Hanrahan}
\author[1]{Francisco Vasconcelos}
\author[3]{You Pang}
\author[3]{Zhen Chen}
\author[3]{Jinlin Wu}
\author[4]{Xiaoyang Zou}
\author[4]{Guoyan Zheng}
\author[5]{Abdul Qayyum}
\author[6]{Moona Mazher}
\author[7]{Imran Razzak}
\author[8]{Tianbin Li}
\author[8]{Jin Ye}
\author[8]{Junjun He}
\author[9,10,11]{Szymon Płotka}
\author[9]{Joanna Kaleta}
\author[12]{Amine Yamlahi}
\author[12]{Antoine Jund}
\author[12,13,14]{Patrick Godau}
\author[15]{Satoshi Kondo}
\author[16]{Satoshi Kasai}
\author[17]{Kousuke Hirasawa}
\author[18,19]{Dominik Rivoir}
\author[20]{Alejandra Pérez}
\author[20]{Santiago Rodriguez}
\author[20]{Pablo Arbeláez}
\author[1,*]{Danail Stoyanov}
\author[1,2,*]{Hani J. Marcus}
\author[1,*]{Sophia Bano}

\affil[1]{Wellcome/EPSRC Centre for Interventional and Surgical Sciences, University College London, London, UK}
\affil[2]{Department of Neurosurgery, National Hospital for Neurology and Neurosurgery, London, UK}
\affil[3]{Centre for AI and Robotics (CAIR) HKISI, CAS, Hong Kong, China}
\affil[4]{Institute of Medical Robotics, School of Biomedical Engineering, Shanghai Jiao Tong University, Shanghai, China}
\affil[5]{National Heart and Lung Institute, Faculty of Medicine, Imperial College London, UK}
\affil[6]{Centre for Medical Image Computing, University College London, London, UK}
\affil[7]{University of New South Wales, Sydney, Australia}
\affil[8]{Shanghai AI Lab, Shanghai, China}
\affil[9]{Informatics Institute, University of Amsterdam, Amsterdam, Netherlands}
\affil[10]{Department of Biomedical Engineering and Physics, Amsterdam University Medical Center, University of Amsterdam, Amsterdam, Netherlands}
\affil[11]{Sano Center for Computational Medicine, Krakow, Poland}
\affil[12]{German Cancer Research Center (DKFZ) Heidelberg, Division of Intelligent Medical Systems, Germany}
\affil[13]{National Center for Tumor Diseases (NCT), NCT Heidelberg, a partnership between DKFZ and University Hospital Heidelberg, Heidelberg, Germany}
\affil[14]{Faculty of Mathematics and Computer Science, Heidelberg University, Heidelberg, Germany}
\affil[15]{Muroran Institute of Technology, Hokkaido, Japan}
\affil[16]{Niigata University of Health and Welfare, Niigata, Japan}
\affil[17]{Konica Minolta Inc., Osaka, Japan}
\affil[18]{National Center for Tumor Diseases, Dresden, Germany: DKFZ, UKDD, TUD, HZDR}
\affil[19]{Centre for Tactile Internet, TUD, Dresden, Germany}
\affil[20]{Universidad de los Andes, Bogota, Colombia}
\affil[*]{These authors contributed equally as senior authors.}
\affil[+]{adrito.das.20@ucl.ac.uk}

\begin{document}
\twocolumn[
\begin{@twocolumnfalse}
\vspace{-4cm}
\maketitle
\thispagestyle{empty}
\begin{abstract} \noindent
The field of computer vision applied to videos of minimally invasive surgery is ever-growing. Workflow recognition pertains to the automated recognition of various aspects of a surgery: including which surgical steps are performed; and which surgical instruments are used. This information can later be used to assist clinicians when learning the surgery; during live surgery; and when writing operation notes. The Pituitary Vision (PitVis) 2023 Challenge tasks the community to step and instrument recognition in videos of endoscopic pituitary surgery. This is a unique task when compared to other minimally invasive surgeries due to the smaller working space, which limits and distorts vision; and higher frequency of instrument and step switching, which requires more precise model predictions. Participants were provided with 25-videos, with results presented at the MICCAI-2023 conference as part of the Endoscopic Vision 2023 Challenge in Vancouver, Canada, on 08-Oct-2023. There were 18-submissions from 9-teams across 6-countries, using a variety of deep learning models. A commonality between the top performing models was incorporating spatio-temporal and multi-task methods, with greater than $50\%$ and $10\%$ macro-\F{} improvement over purely spacial single-task models in step and instrument recognition respectively. The PitVis-2023 Challenge therefore demonstrates state-of-the-art computer vision models in minimally invasive surgery are transferable to a new dataset, with surgery specific techniques used to enhance performance, progressing the field further. Benchmark results are provided in the paper, and the dataset is publicly available at: \url{https://doi.org/10.5522/04/26531686}.  

\smallbreak \noindent
\end{abstract}
\end{@twocolumnfalse}
]

\section{Introduction} \label{sec:intro}

\begin{figure}[t]
\centering
\includegraphics[width=0.95\columnwidth]{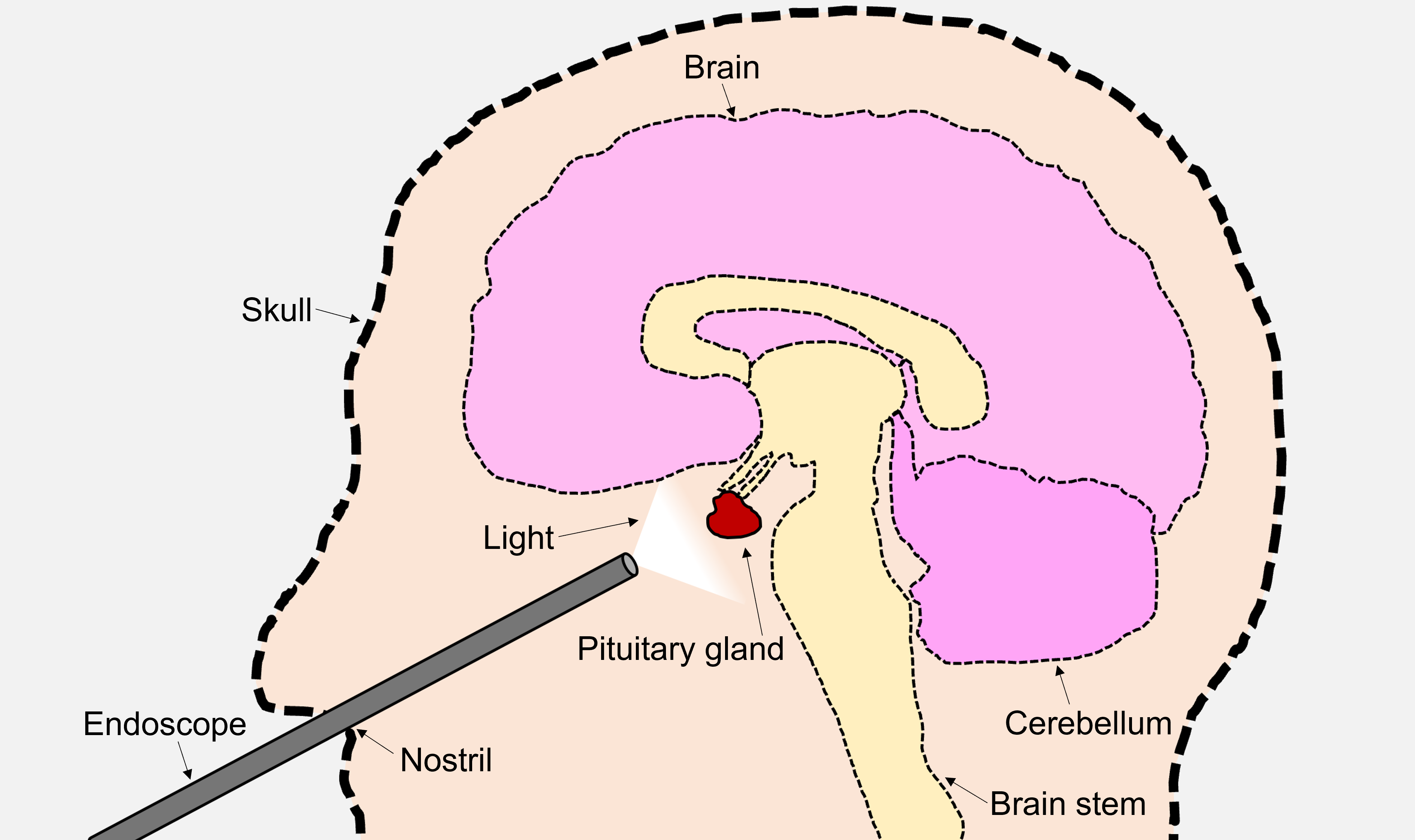}
\caption{Endoscopic pituitary surgery diagram.}
\label{fig:01pituitary_diagram}
\end{figure}

The \gls{EV} challenge\footnote{\url{https://opencas.dkfz.de/endovis/}} has existed since 2015, hosted by the \gls{MICCAI} Society \cite{MaierHein2022}. Included are a wide range of challenges related to computer vision in minimally invasive surgeries: from polyp detection in colonoscopy videos in 2015 to action recognition on radical prostatectomy videos in 2022 \cite{MaierHein2022}. To the minimally invasive surgical computer vision community, the benefits of an \gls{EV} challenge are two-fold: (i) it pushes the boundaries of existing models \cite{MaierHein2022}; and (ii) it provides a curated public dataset \cite{MaierHein2022}. Building on this, the \gls{PV} 2023 challenge was created as sub-challenge of \gls{EV}-2023 \cite{EndoVis2023}. The \gls{PV}-2023 challenge pertains to step and instrument recognition in the \gls{eTSA} for pituitary adenoma resection.

The pituitary gland is found at the base of the brain \cite{Ganapathy2022}. Tumours of the anterior pituitary gland, pituitary adenomas, have an estimated prevalence of 1 in 1000 of the general population \cite{Russ2022, Agustsson2015}. Symptoms typically include visual impairment \cite{Russ2022,Ogra2014} and hormone imbalances \cite{Ganapathy2022, Russ2022}. Left untreated, these symptomatic adenomas can cause blindness \cite{Russ2022,Ogra2014} or, in cases such as Cushing’s disease, be life limiting \cite{Russ2022,Tritos2019}. The gold standard treatment for most patients with a symptomatic pituitary adenoma is surgery, commonly via the \gls{eTSA} \cite{Ganapathy2022,Wang2014}.

\begin{figure}[t]
\centering
\includegraphics[width=0.95\columnwidth]{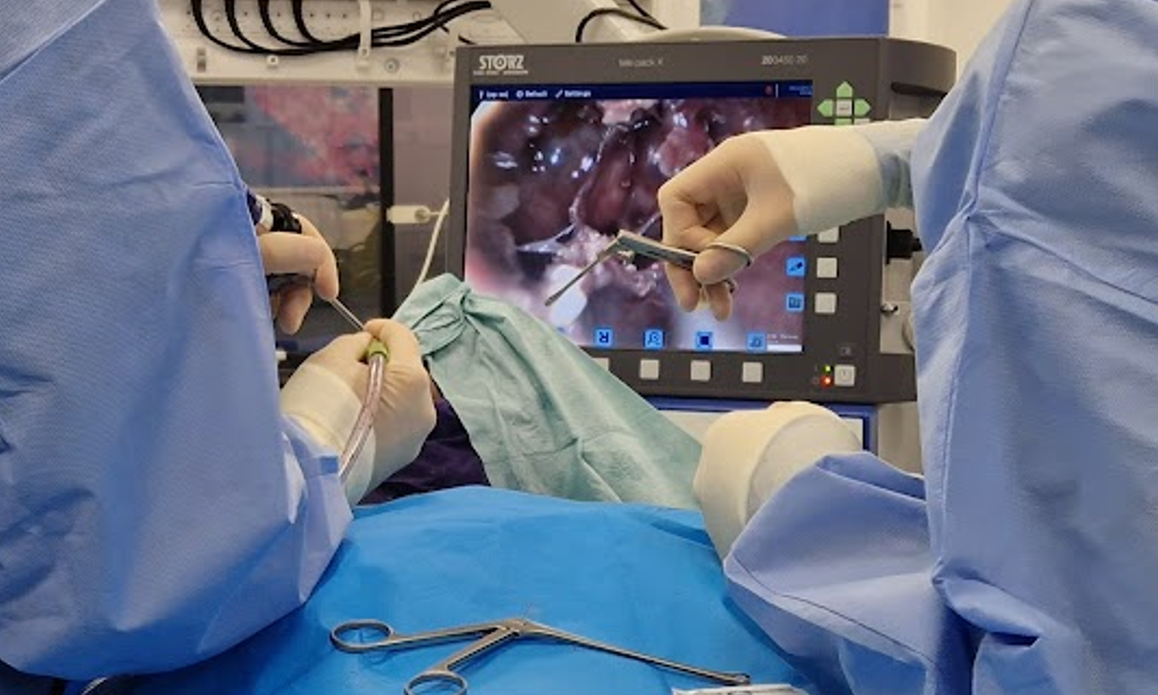}
\caption{Endoscopic pituitary surgery operation.}
\label{fig:02pituitary_operation}
\end{figure}

The \gls{eTSA}, also called endoscopic pituitary surgery, is a minimally invasive surgery where the tumour is removed by entering through a nostril, as displayed in \fig{fig:01pituitary_diagram} \cite{Wang2014,Marcus2021}. The endoscope allows the surgeon to see inside the patient, with the camera feed projected onto a monitor, and is used in conjunction with surgical instruments, as displayed in \fig{fig:02pituitary_operation} \cite{Wang2014,Marcus2021}. The \gls{eTSA} is performed heterogeneously \cite{Cranial2023}, and so there is variability in outcomes \cite{Wang2014}. Furthermore, it is a difficult procedure to master, requiring dedicated sub-specialty training \cite{Frara2020}.

The \gls{eTSA} can be broken down into granular clinical steps, using various instruments to achieve the task of a given step \cite{Marcus2021}. Workflow recognition is the name given to the automated recognition of these steps and instruments \cite{Marcus2021,Wang2022}, and can aid clinicians in a variety of ways, including: (i) Teaching junior surgeons via interactive videos and coaching via automated performance metrics, and hence reducing the steep learning curve \cite{Khan2022, Khan2023, Khan2024B}. (ii) After a surgery, by automating the reporting of steps performed and instruments used, which will reduce the time spent on the writing of operation notes \cite{Khan2023, Das2023, He2024}. (iii) During live surgery, automatically informing the wider operating room team (e.g. anaesthetists and theatre nurses) when a new step is to begin or when a new instrument is required, in order to improve operating room efficiency \cite{Khan2023, Khan2024A, Garrow2020}.

\newpage

Motivated by these clinical benefits, the \gls{PV}-2023 challenge was created. The challenge consisted of three tasks: (1) step recognition; (2) instrument recognition; and (3) step and instrument recognition. Participants were provided with 25-training-videos (public), along with per-second annotations of the current step and present instrument. Submitted models were evaluated on 8-testing-videos (private), and monetary prizes totalling £3000 were awarded. The main contributions of the \gls{PV}-2023 challenge are as follows:

\begin{enumerate} 
\itemsep-0.5em 
\item A thorough analysis of the state-of-the-art surgical workflow recognition models applied to endoscopic pituitary surgery: more granular than previous step recognition work and the first for instrument recognition in this surgery. \item Providing benchmark results of surgical workflow recognition in endoscopic pituitary surgery, highlighting the challenges on a unique surgery not previously explored by the community. 
\item The first curated public dataset of endoscopic pituitary surgery: 25-videos with each second annotated with its respective step and instrument.
\item A well-attended computer vision challenge associated with endoscopic pituitary surgery: with 18-submissions from 9-teams across 6-countries.
\end{enumerate}

This paper follows the BIAS guidelines for transparent reporting of biomedical challenges \cite{MaierHein2020}.

\section{Related works} \label{sec:works}
\subsection{Difficulties}
In minimally invasive surgery, workflow recognition is a difficult computer vision task for several reasons, including: (i) A variety in surgical practice across different hospitals throughout the globe, resulting in a lack of consensus of which steps are to be performed and instruments to be used \cite{Garrow2020, Rueckert2024}. (ii) A limited supply of well-curated large annotated public datasets, resulting in models focusing on some surgeries (e.g. laparoscopic cholecystectomy) and so their generalisability has not been well studied \cite{Wang2022, Demir2023}.
(iii) Poor metric selection, often not representative of the underlying clinical motivation \cite{Wang2022, MaierHein2024}. 

Additionally, there are several \gls{eTSA} specific difficulties, including: (iv) Multiple steps and instruments with a high frequency of switching in an undetermined order, more so than in other surgeries \cite{Marcus2021, Garrow2020, Das2022}. This increases classification difficulty as the model predictions need to be more precise. (v) The small working space, leading to a thinner endoscope, and hence lense distortion \cite{Das2022}. This means features at the center of the image appear smaller than features towards the edge of an image. This leads to instrument shafts, which are generally uninformative of the instrument class, to take up a large section of the image; whereas instrument tips, which are more informative of the instrument class, take up a small section of the image (\fig{fig:04instruments_images}). (vi) Occlusions due to bodily fluids, necessitating the need for the frequent withdrawal of the endoscope outside of the patients body for cleaning, resulting in temporally inconsistent images \cite{Das2023, Das2022}. (vii) Many of the steps and instruments look similar. For example, instrument-9 (micro doppler probe) and instrument-18 (tissue glue applicator) look identical from a static image, and can only be distinguished by the action performed and the wider surgical context (\fig{fig:04instruments_images}).

\subsection{Step recognition} 
Historically, a variety of machine learning models were used for step recognition across minimally invasive surgeries, but since 2016, deep learning models have dominated \cite{Garrow2020, Demir2023}. Typically, step recognition models consist of a 3-stage architecture: stage-1, a per-frame spatial encoder; followed by stage-2, where the per-frame spatial features are consecutively combined and sent to a temporal decoder; and finally stage-3, where the predicted spatial-temporal classifications are turned into a sequence and undergo processing \cite{Garrow2020, Demir2023}. For stage 1, \glspl{CNN} are frequently used, although more recently \glspl{S-TF}  transformers or \glspl{ST-TF} have been found to be effective \cite{Demir2023}. For stage 2, \glspl{TCN}; \glspl{T-TF}; and \glspl{RNN} often used, particularly \glspl{LSTM} and \glspl{GRU} \cite{Garrow2020, Demir2023}. For stage 3, \glspl{HMM} were typically used \cite{Garrow2020, Demir2023, Twinanda2017}, but other methods, such as \glspl{TSF}, are also common \cite{Das2022}.

For the \gls{eTSA}, a \gls{CNN} + \gls{LSTM} + \gls{TSF} architecture was shown to be the best performing \cite{Das2022}. More specifically, ResNet50 was used as the spatial feature extractor, and the 10-frames feature output was fed into an \gls{LSTM}, before a threshold smoothing function was used \cite{Das2022}. The smoothing function ensured the step predictions were consistent for a certain period of time before switching to another step, to reduce the number of the frequent yet short periods of incorrect predictions \cite{Das2022}. The model was trained on 40-videos and validated on 10-videos, achieving a 0.74 weighted-\F{} in 7-step frame-level classification (5-fold-cross-validation) \cite{Das2022}. Based on this model, a \gls{CNN} + \gls{TSF} architecture was used to predict the presence of a step in a given video of \gls{eTSA} to then automatically generate the usually manually written operation notes \cite{Das2023}. In this more recent work, the model was trained on 77-videos and tested on 20-videos, achieving a 0.80 weighted-\F{} in 27-step multi-label video-level classification \cite{Das2023}. 

\subsection{Instrument recognition}
The majority of computer vision models created for minimally invasive surgeries regarding instruments is to accomplish instrument segmentation, rather than instrument recognition \cite{Wang2022, Rueckert2024}. Instrument segmentation is an extension of instrument recognition, where the type of instrument needs to not only be classified (instrument recognition) but the boundaries of the instrument also needs to be predicted. Due to this more difficult task, more sophisticated models, utilising an encoder-decoder architecture are used. However, similar to step recognition models, the most common encoders are \glspl{CNN} for spatial feature extraction and \glspl{RNN} for temporal feature extraction \cite{Wang2022, Rueckert2024}. No work has been published for instrument recognition for the \gls{eTSA}. 

\subsection{Multi-task recognition}
Multi-task step and instrument recognition models connect single-task models at various stages in the neural network architecture \cite{Twinanda2017, Psychogyios2024, Oluwatosin2024}. In doing so, they outperform single-task models in both tasks by sharing information \cite{Das2023B, Mao2024}. For example, in \cite{Jin2020}, a joint spatial-temporal (\gls{CNN} + \gls{RNN}) backbone is used for feature extraction in combination with a correlation loss function, so information gained from one task is shared with the other. However, multi-task models are not commonly used due to a lack of data \cite{Wang2022, Oluwatosin2024}. No work has been published for multi-task step and instrument recognition for the \gls{eTSA}. 

\section{Challenge description} \label{sec:challenge}
The aim of the \gls{PV}-2023 challenge was to develop \gls{ML} models capable of step and instrument recognition in the \gls{eTSA}. In doing so, these models provide surgical context that can be used as an assistive tool for clinicians. 
\subsection{Tasks}

\begin{figure*}[!t]
\centering
\includegraphics[width=0.87\textwidth]{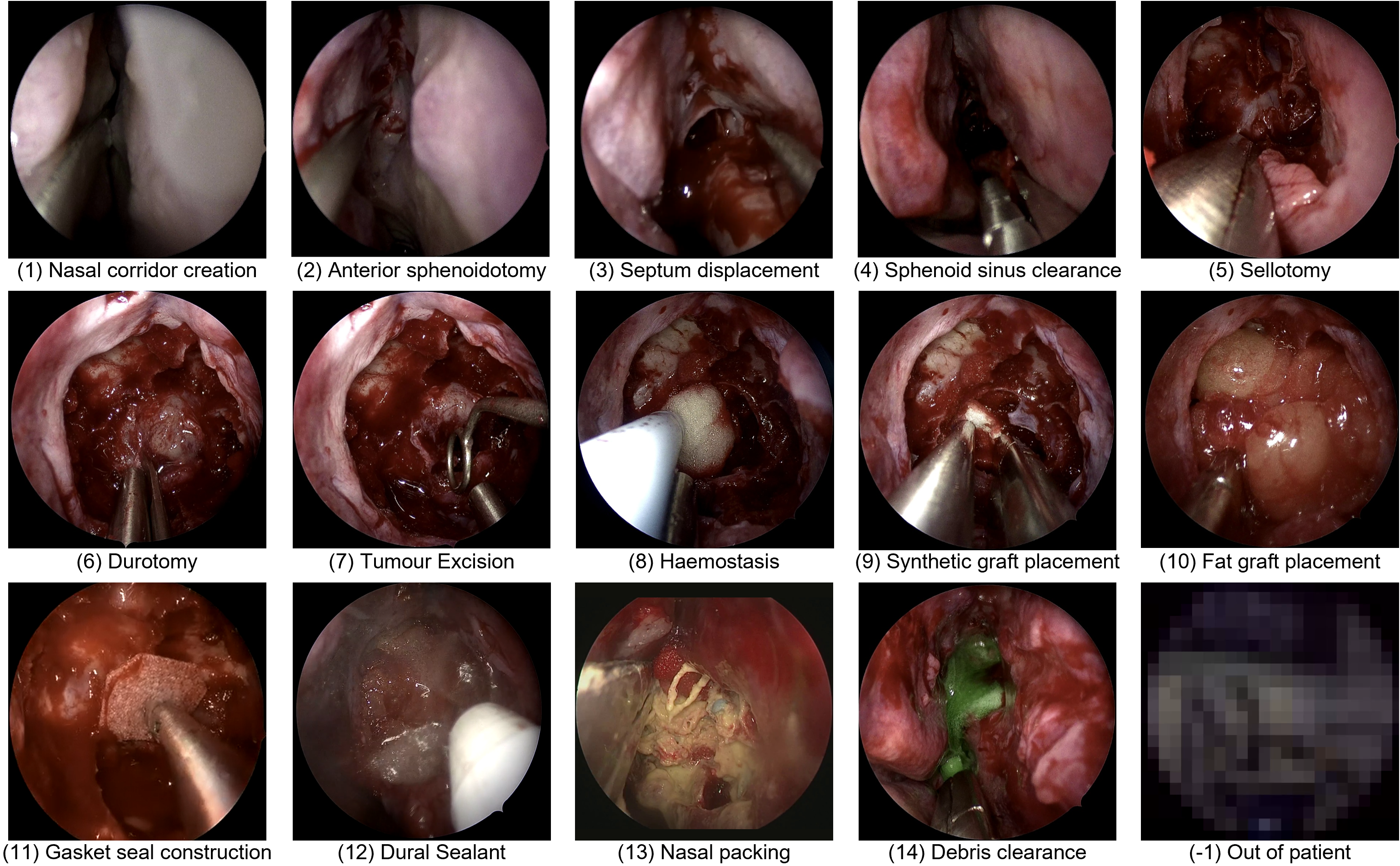}
\caption{Representative images of each of the 14-steps. Note step-11 and step-13 were not evaluated due to having insufficient volume to train on (\fig{fig:07steps_distribution}), and \Q{out of patient} is not considered a class.}
\label{fig:03steps_images}
\end{figure*}

\begin{figure*}[!t]
\centering
\includegraphics[width=0.93\textwidth]{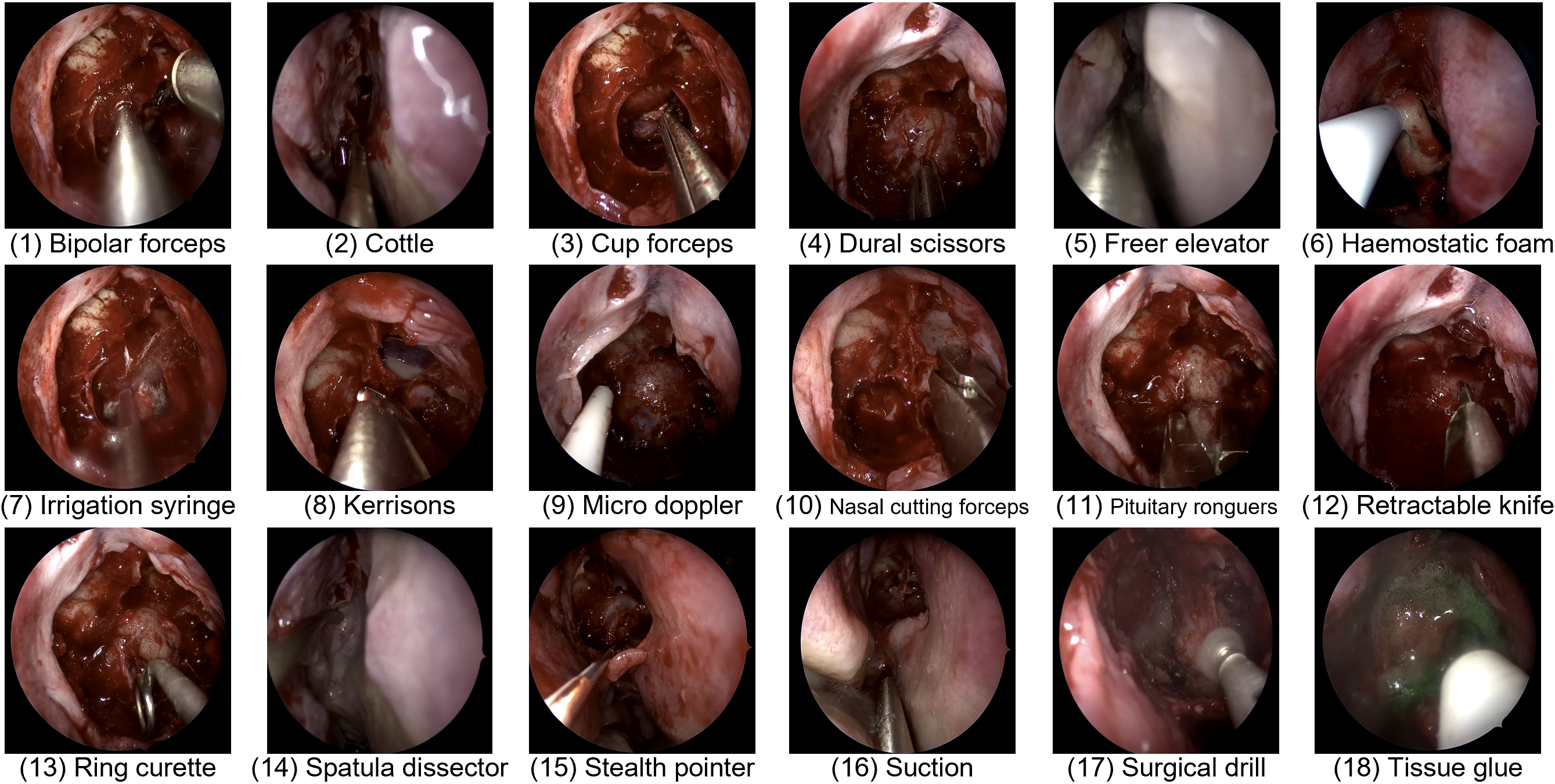}
\caption{Representative images of each of the 18-instruments, excluding the \Q{no instrument} class.}
\label{fig:04instruments_images}
\end{figure*}

\begin{figure*}[!t]
\centering
\includegraphics[width=\textwidth]{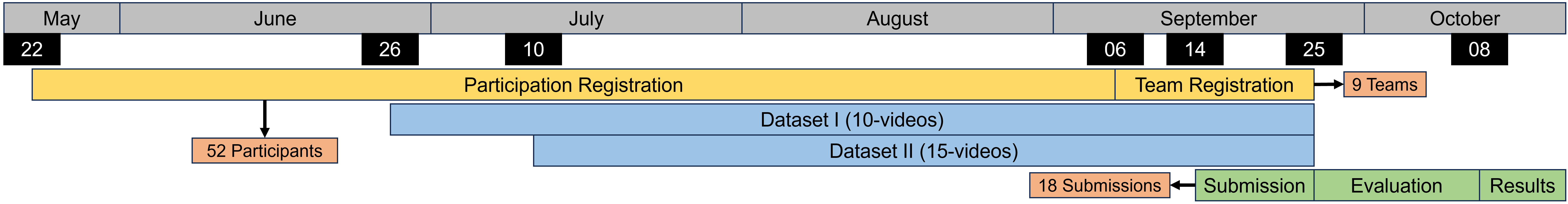}
\caption{A timeline of the challenge. All dates are in 2023.}
\label{fig:05timeline}
\end{figure*}

The challenge consisted of 3-tasks:

\begin{enumerate} 
\itemsep0em 
\item Surgical step recognition.
\item Surgical instrument recognition.
\item Multi-task steps and instrument recognition.
\end{enumerate}

Representative images of the 12-steps and 19-instruments are displayed in \fig{fig:03steps_images} and \fig{fig:04instruments_images} respectively. These steps and instruments are defined in \cite{Marcus2021}, and confirmed by two neurosurgical trainees (DZK and JGH) and one consultant neurosurgeon (HJM), based on the training dataset. For task-1; exactly one step is present at a given time, hence this is a multi-class problem. For task-2; zero, one, or two instruments may be present at a given time, hence this is a multi-label problem. Task-3 is a combination of task-1 and task-2, hence a multi-task problem.

\subsection{Organisation}
The \gls{PV}-2023 challenge was a one-time event as part of \gls{EV}-2023 \cite{EndoVis2023}, with all results presented publicly at the \gls{MICCAI}-2023 conference in Vancouver, Canada. A timeline of the challenge organisation is displayed in \fig{fig:05timeline}. Organisation, communication, data sharing, and submissions were all done via the Synapse challenge website\footnote{\url{www.synapse.org/\#!Synapse:syn51232283/wiki/621581}}, and no private communication with the organisers was permitted.

The organisation committee consisted of a collaboration between computer scientists and neurosurgeons from the \gls{WEISS} at \gls{UCL}, London, \gls{UK} and the Department of Neurosurgery at the \gls{NHNN}, London, \gls{UK} respectively. 

Advertisement was predominately done via social media\footnote{\url{www.x.com/AdritoDas/status/1660677465956548609}}. 52-participants registered to download the data, with 9-teams across 6-countries successfully submitting 18-submissions. Prizes totalling £1000 per task were available to the top-2 teams of each task. Teams from \gls{WEISS} were allowed to submit models, but illegible to win prizes.

25-annotated-videos were provided. A 20-training to 5-validation (01, 12, 21, 24, 25) split was suggested but not enforced. This split was based on step and instrument distributions (\secr{sec:dist}), such that the number of annotations for a class remained at an approximate 4:1 ratio, as is common in workflow recognition \cite{Rueckert2024, Demir2023}. The 8-testing-videos were not provided to the participants. The training and testing videos are similar to those of the intended use cases. 

\subsection{Model requirements}
Only fully-automatic methods were permitted: the model must have taken an image input and output the predicted classification(s) as appropriate for the given task. For task-3, a multi-task model is defined as a single model that takes an image input and outputs both a predicted step classification and a predicted instrument classification congruently. 

Only online models were permitted: only information from frames up to and including the current frame can be used to classify the current frame.

Using instrument annotations for step recognition training, or using step annotations for instrument recognition training was permissible. Training on publicly available data was permissible if stated in the participant's submission description. 

Models were submitted as docker containers via Synapse on the challenge website, after detailed submission instructions were given. This included an example docker submission with the associated evaluation scripts, downloadable from GitHub\footnote{\url{www.github.com/dreets/pitvis/}}. The status of whether a submission was successfully submitted could also be found on the challenge website, but not the final evaluation scores. Participants were not required to publish their code, but were required to give detailed descriptions and diagrams of their model. Finalised dockers were run on on single core of an NVIDIA-Tesla-V100-Tensor-Core-32-GB-GPU, and had to run in a reasonable time (less than 1 minute of runtime for every 10 minutes of video). 

\clearpage

\subsection{Evaluation metrics}
\subsubsection{Spatial metric}
Macro-\F{} (\eq{eq:mf1}) was the chosen spatial metric. This is because \F{} (\eq{eq:f1}) ensures a high per-frame accuracy while also safeguarding against small precision or recall. Taking a macro-mean across classes ensures each class is treated equally so major classes do not dominate.

\begin{equation} \text{Macro-\F{}}=\frac{1}{N}\sum_{i=1}^{N} (\F{})_{i} {\ ,} \label{eq:mf1} \end{equation}
\begin{equation} \F{}=\frac{2\text{TP}}{2\text{TP} + \text{FP} + \text{FN}} {\ ,} \label{eq:f1} \end{equation}
\noindent where $N$ = total number of classes; TP = true positive; FP = false positive; FN = false negative. 

\subsubsection{Temporal metric}
\E{} (\eq{eq:edit}) was chosen as the temporal metric \cite{Lea2016}. It is the reciprocal of the Leveshtein distance (\eq{eq:Lev}), which measures the number of edits (insertions, deletions, substitutions) required to change one temporal series into the other, and by doing so, penalises temporally inconsistent predictions \cite{Lea2016}. A series is defined as classifications without repeats. For example, classifications $[0, 0, 0, 1, 1, 0, 1, 1]$ are compressed to a $[0, 1, 0, 1]$ series.

\begin{equation}\E{} = \frac{1}{\text{Lev}} {\ ,}\label{eq:edit}\end{equation}
\begin{multline}\noindent \text{Lev}(s, t) = \\ \begin{cases} |s| & \text{if } |t| = 0, \\ |t| & \text{if } |s| = 0, \\ \text{Lev}\big(\text{tail}(s),\text{tail}(t)\big) & \text{if } \text{head}(s) \\ & = \text{head}(t), \\ 1 + \min \begin{cases} \text{Lev}\big(\text{tail}(s), t\big) \\ \text{Lev}\big(s, \text{tail}(t)\big) \\ \text{Lev}\big(\text{tail}(s), \text{tail}(t)\big) \\ \end{cases} & \text{otherwise.} \end{cases}{\ ,}\label{eq:Lev}\end{multline}
\noindent where $\text{head}(s)$ is the first value; and $\text{tail}(s)$ is all but the first value of a given series $s$.

\subsubsection{Task specific metrics}
The mean of Macro-\F{} and \E{} was chosen as the step recognition metric (\eq{eq:task1}). This is so models are optimised for both frame-level accuracy and temporal consistency. Previous work has shown using purely spatial metrics leads to a high \F{} but frequent inaccurate changes of steps for short periods of time \cite{Das2022}. 

\begin{equation} \frac{\text{12-steps-Macro-\F{}} + \text{12-steps-\E{}}}{2} \label{eq:task1} \end{equation}

Macro-\F{} was the chosen metric for instrument recognition with no \E{} (\eq{eq:task2}). This was because the usage of instruments is much more volatile and heavily dominated by the instrument-0 (no instrument) and instrument-16 (suction) class (\fig{fig:09instruments_distribution}). For example, a typical snippet of a ground-truth sequence is $[0, 11, 0, 0, 11, 16, 16, 11, 16]$, where an instrument such as instrument-11 (pituitary ronguers) will be briefly used between the dominating instrument-0 and instrument-16 classes. This means an incorrect prediction will be strongly penalised by temporal metrics. Moreover, as instrument recognition is a multi-label problem, a single sequence does not encapsulate all of the data, and so more sophisticated temporal metrics beyond \E{} are required. After the results of this challenge, and the models are analysed, an appropriate temporal metric will be used for future work in an attempt to improve temporal consistency.

\begin{equation} \text{19-instruments-Macro-\F{}} \label{eq:task2}\end{equation}

The mean-average of the respective step and instrument recognition metric was chosen as the multi-task metric (\eq{eq:task3}). This was done to treat both step and instrument recognition equally.
\begin{equation} \frac{\eq{eq:task1} + \eq{eq:task2}}{2} \label{eq:task3}\end{equation}

\newpage

\section{Dataset} \label{sec:dataset}
The challenge dataset is the first publicly available annotated dataset of the \gls{eTSA}. This section describes the dataset acquisition and analyses its properties. 

\subsection{Data acquisition}
\subsubsection{Videos}
The \gls{NHNN} (Queens Square, London, \gls{UK}) provided all videos used in the \gls{PV} challenge. This hospital is an academic tertiary neurosurgical centre, performing 150-200 pituitary operations each year \cite{Khan2022}. Videos of the \gls{eTSA} were excluded if: the operation was a revision surgery within 6-months of the primary surgery; if large sections of the surgery were missing; or if the surgery was significantly different from a usual surgery. This curation was performed by two trainee neurosurgeons (DZK and JGH) and verified by a consultant neurosurgeon (HJM). The dataset size was determined by what was feasible to annotate in the challenge timeline. 

The 25-training-videos were recorded between 02-Jul-2021 and 28-Dec-2022, and have written consent for public research use. The 8-testing-videos were recorded between 06-Dec-2018 to 07-Jan-2021, and have consent for research use within the organisers' institute (\gls{UCL}). The study was registered with the \gls{UCL} \gls{IRB} (17819/011).

The surgeries were recorded using a high-definition endoscope (Hopkins Telescope with AIDA storage system, Karl Storz Endoscopy\footnote{\url{www.karlstorz.com/}}, \gls{UK}). The original videos have a variable \gls{fps}, with resolutions varying from 720p-2160p. These videos were uploaded from the hospital servers to the commercially available Touch Surgery\textsuperscript{TM} Ecosystem\footnote{\url{www.touchsurgery.com/}}, an AI-powered surgical video management and analytics platform provided by Medtronic. Here, the videos were de-identified by blurring all images outside of the patient. The videos were then converted to a constant 24-\gls{fps} with 720p resolution using the publicly available Handbrake\footnote{\url{www.handbrake.fr/}}, and stored as .mp4 files.

Additionally, a script to sample the videos at 1 \gls{fps}, and store them as .png images was provided on the GitHub. This sampling script was used by the organisers on the 8-testing-videos, and the .png images were fed into the submitted models for evaluation. 

\subsubsection{Annotations}
\begin{table}[!t]
\centering
\resizebox{\columnwidth}{!}{
\begin{tabular}{c|c|c|c|c}
int\_video & int\_time & int\_step & int\_instrument1 & int\_instrument2 \\ \hline
25 & 0 & -1 & -1 & -2 \\
25 & 1 & -1 & -1 & -2 \\
… & … & … & … & … \\
25 & 2011 & 5 & 8 & 16 \\
25 & 2012 & 5 & 16 & -2 \\
25 & 2013 & 5 & 16 & -2 \\
25 & 2014 & 5 & 0 & -2
\end{tabular}
}
\caption{An example of the .csv annotations given to participants. A \Q{-2} in the \Q{int\_instrument2} column is indicative of \Q{no annotation}. Note \Q{...} indicates a break in the annotations for demonstration purposes.}
\label{tab:02annotations}
\end{table}

For steps, each video was annotated by two trainee neurosurgeons (DZK and JGH)  with any discrepancies solved via discussion and mutual agreement. For instruments, a third-party company Anolytics\footnote{\url{www.anolytics.ai/}} was used. These annotations were not performed by clinical specialists, but verified by one neurosurgical trainee (DZK) and one research scientist (AD). All annotations were then verified by a consultant neurosurgeon (HJM) before being released.

Annotations were released as .csv files along with their associated videos, an example of which is displayed in \tab{tab:02annotations}. The map of the step or instrument to the corresponding integer was also provided. 

As with all annotations, there can be errors, and in this challenge the most likely source is human error in misidentifying a step or instrument. These were reduced by the aforementioned multiple rounds of annotating and verification, and if any were found after release, they were immediately corrected and participants were informed.

\subsection{Data analysis} \label{sec:dist}

\begin{figure}[!t]
\centering
\includegraphics[width=\columnwidth]{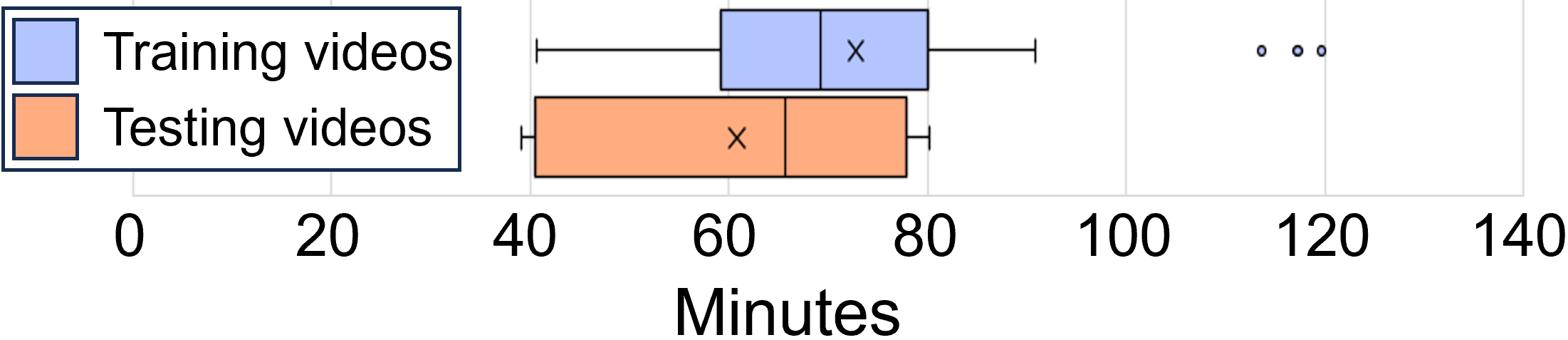}
\caption{Length distribution of the 25-training and 8-testing videos without the \Q{out of patient} class.}
\label{fig:06videos_distribution}
\end{figure}

\begin{figure}[!t]
\centering
\includegraphics[width=\columnwidth]{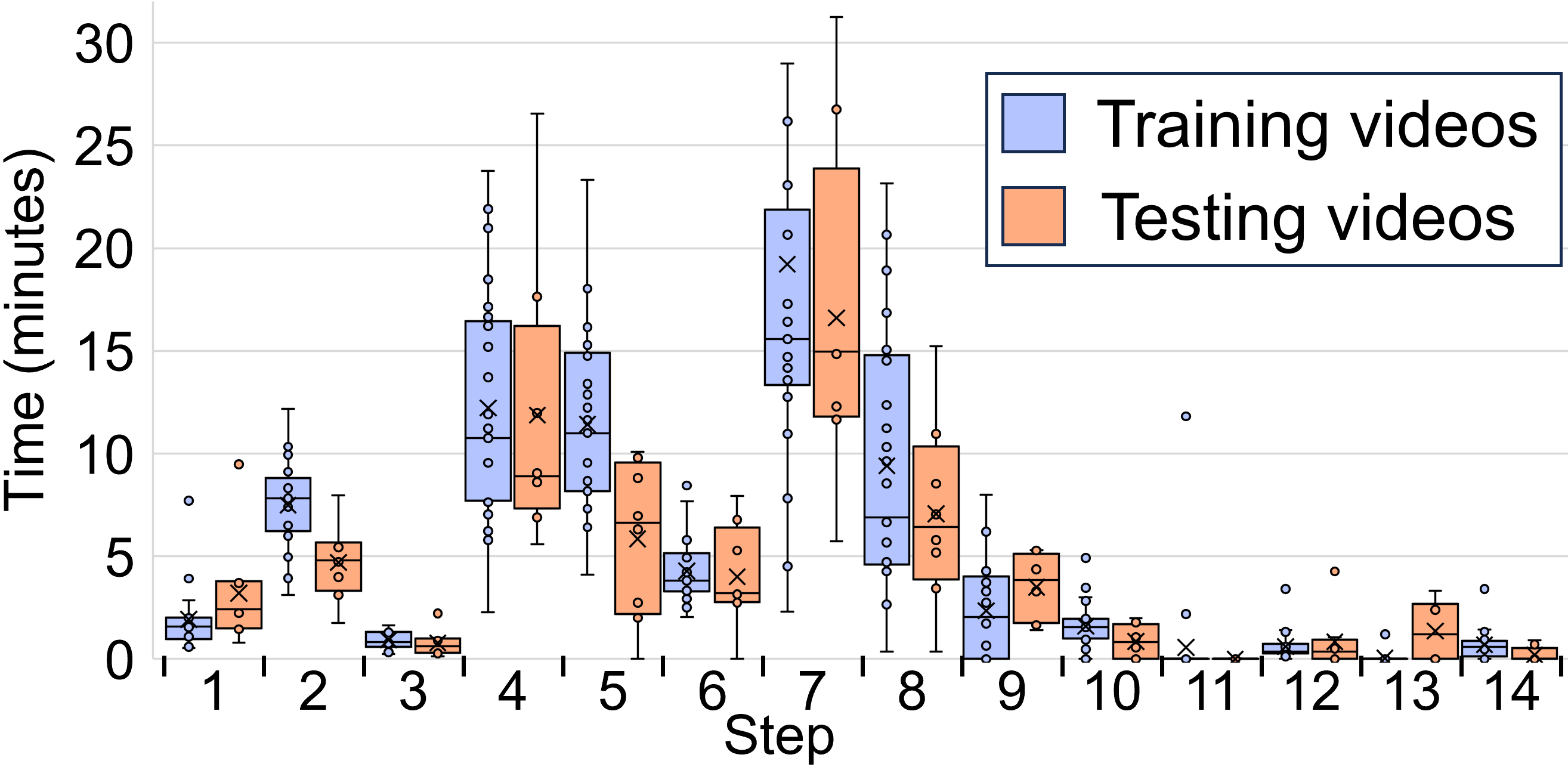}
\caption{Length distribution of steps across the 25-training and 8-testing videos.}
\label{fig:07steps_distribution}
\end{figure}

\begin{figure}[!t]
\centering
\includegraphics[width=\columnwidth]{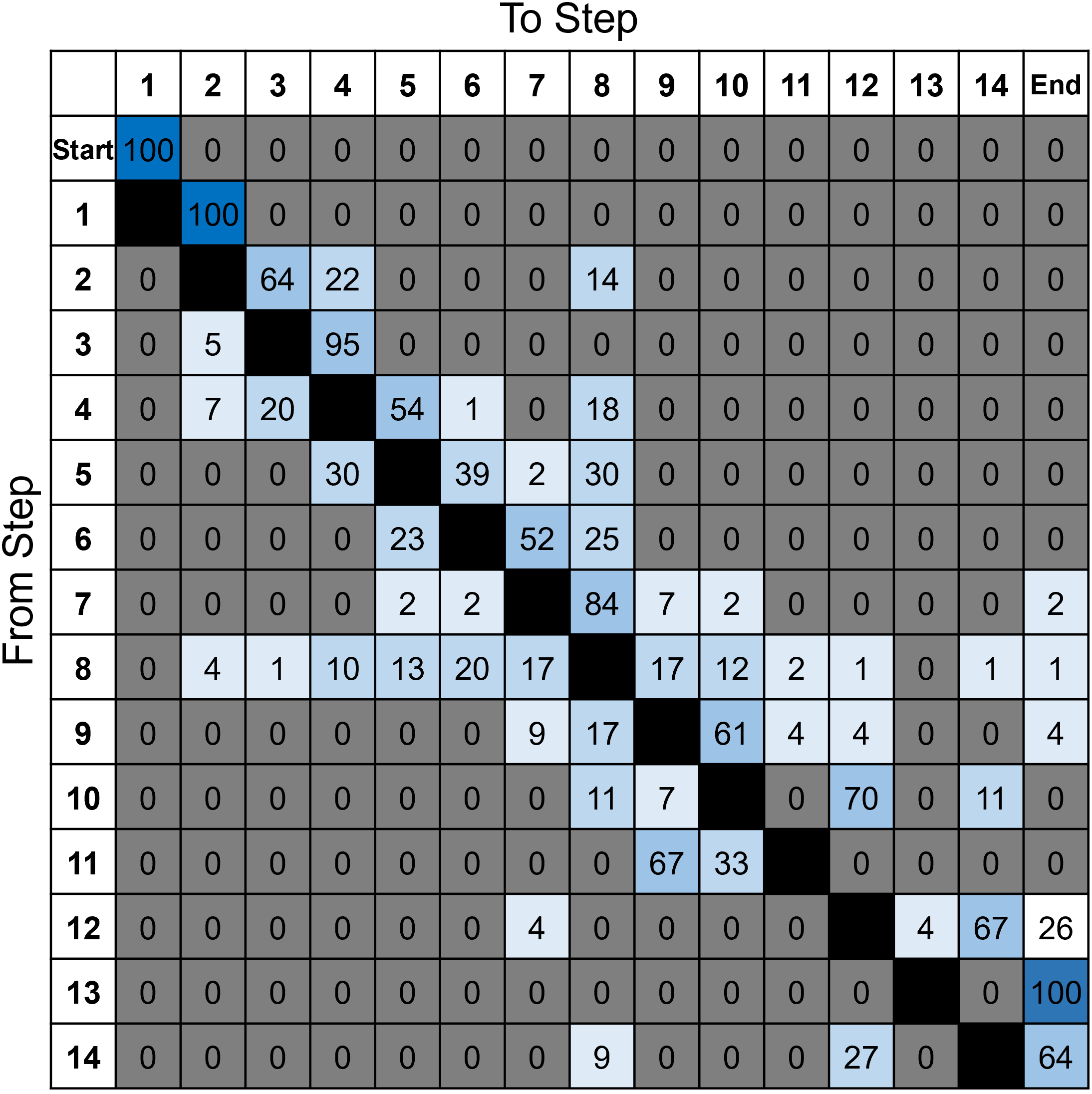}
\caption{Transition probabilities across the 25-training-videos. Each value represents the probability of going from one step to another (e.g. step-4 goes to step-5 with 54\% probability). The \Q{out of patient} class was removed for these calculations.}
\label{fig:08steps_transition}
\end{figure}

\begin{figure*}[!t]
\centering
\includegraphics[width=\textwidth]{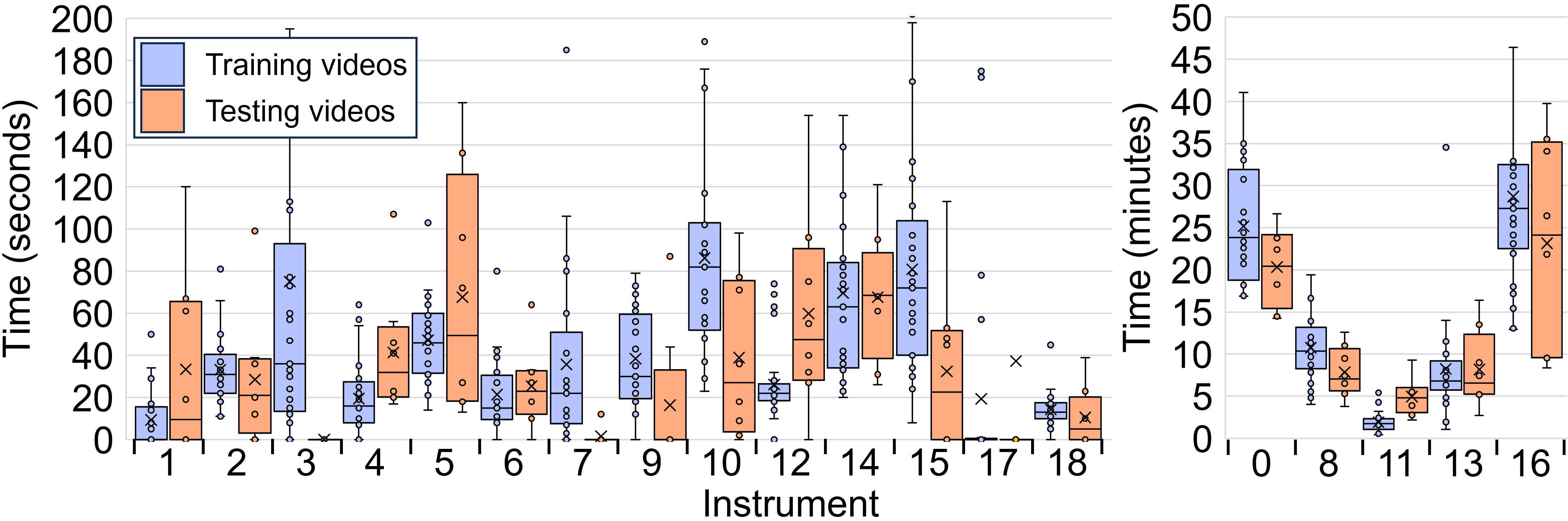}
\caption{Length distribution of instruments across the 25-training and 8-testing videos. The time axis is presented as seconds in the left diagram and minutes in the right diagram - this is for improved visibility, as otherwise the minor class instrument length distributions would not be visible.}
\label{fig:09instruments_distribution}
\end{figure*}

\subsubsection{Videos}
The distribution of video lengths across all videos is displayed in \fig{fig:06videos_distribution}. The mean and median of the 25-training-videos was 72.8+7.2 and 69.2+6.4 minutes respectively, where $+t$ indicates time, $t$, outside of the patient. This was slightly longer than the mean and median of the 8-testing-videos, which were 60.9+5.6 and 65.7+5.3 minutes respectively. 
The \Q{out of patient} frames, indicated by the \Q{-1} class in annotation files were excluded during evaluation. 

\subsubsection{Steps}
Step-11 (gasket seal construct) and step-13 (nasal packing) were only present in 2 and 1 training-videos respectively, and so were removed due to having insufficient volume to train on (\fig{fig:07steps_distribution}), and any such frames were excluded during evaluation. A hypothetical step-0 (no step) class does not exist as every part of a video belongs to a step. 

Steps 1-8 are present in all 25-training-videos, with the remaining steps found in at least 18-training-videos. As displayed in \fig{fig:07steps_distribution}, the length of steps are similar across the training and testing videos, but the step lengths themselves are varied. For example, step-7 (tumour excision) is the longest and step-14 (debris debulking) is the shortest with a with mean lengths of 19.2 and 0.7 minutes respectively. Moreover, as displayed in \fig{fig:08steps_transition}, the transition probabilities from one step to the next are not consistent. For example, step-8 (haemostasis) is often transitioned to and from out of sequence due to its short but frequent occurrences during surgery. This lack of consistency highlights the difficulty of step recognition in this dataset and the \gls{eTSA} in general.

\newpage

\subsubsection{Instruments}
A \Q{-1} annotation indicates the \Q{out of patient} class and \Q{-2} indicates a \Q{no secondary instrument} as to not have an empty entry in this column, and these frames were excluded during evaluation.

The majority of instruments are found in 20 or more training-videos. Exceptions to this are instrument-1 (bipolar forceps), found in 12-videos; and instrument-17 (surgical drill), found in 6-videos. As displayed in \fig{fig:09instruments_distribution}, the length distribution for instruments is dominated by instrument-0 (no instrument)  and instrument-16 (suction) with mean lengths of 25.2 and 28.7-minutes respectively. The remaining instrument lengths are more clustered, although there is still some variance. There are also quite drastic differences between the training and testing dataset. For example, instrument-3 (cup forceps) and instrument-7 (irrigation syringe) have a relatively high usage in the training-videos, but very low usage in the testing-videos. This is likely due to time difference between when the training and testing surgeries were performed: leading to different availability of instruments, and variance in surgical technique. Similar to the steps, this highlights the difficulty of instrument recognition for the \gls{eTSA}.

\newpage

\section{Methods}

\begin{table*}[!t]
\centering
\resizebox{\textwidth}{!}{
\begin{tabular}{|c|c|c|ccc|}
\hline
\multirow{2}{*}{\textbf{Team}} & \multirow{2}{*}{\textbf{Institute}} & \multirow{2}{*}{\textbf{Task}} & \multicolumn{3}{c|}{\textbf{Simplified Model Architecture}} \\
 &  &  & \multicolumn{1}{c|}{\textbf{Stage-1}} & \multicolumn{1}{c|}{\textbf{Stage-2}} & \textbf{Stage-3} \\ \Xhline{5\arrayrulewidth}
\textbf{CAIR-} & Hong Kong Institute of & \multirow{3}{*}{1} & \multicolumn{1}{c|}{\multirow{3}{*}{CSPDarknet53(\gls{CNN})}} & \multicolumn{1}{c|}{\multirow{3}{*}{TeCNO\{10\}(\gls{TCN})}} & \multirow{3}{*}{-} \\
\textbf{POLYU-} & Science and Innovation &  & \multicolumn{1}{c|}{} & \multicolumn{1}{c|}{} &  \\
\textbf{HK} & Hong Kong, China &  & \multicolumn{1}{c|}{} & \multicolumn{1}{c|}{} &  \\ \Xhline{5\arrayrulewidth}
\multirow{3}{*}{\textbf{CITI}} & Shanghai Jiao & \multirow{2}{*}{1,3} & \multicolumn{1}{c|}{\multirow{3}{*}{Swin\{20\}(\gls{ST-TF})}} & \multicolumn{1}{c|}{\multirow{2}{*}{ARST\{80\}(\gls{ST-TF})$\langle$step$\rangle$}} & \multirow{2}{*}{-} \\
 & Tong University & & \multicolumn{1}{c|}{} & \multicolumn{1}{c|}{} &  \\ \cline{3-3} \cline{5-6}
 & Shanghai, China & 2 & \multicolumn{1}{c|}{} & \multicolumn{1}{c|}{-} & - \\ \Xhline{5\arrayrulewidth}
\multirow{2}{*}{\textbf{DOLPHINS}} & Imperial College London & \multirow{2}{*}{1} & \multicolumn{1}{c|}{XCiT(\gls{S-TF})} & \multicolumn{1}{c|}{\multirow{2}{*}{Pairwise Ensemble}} & \multirow{2}{*}{-} \\ \cline{4-4}
 & London, UK &  & \multicolumn{1}{c|}{DenseNet201(\gls{CNN})} & \multicolumn{1}{c|}{} &  \\ \Xhline{5\arrayrulewidth}
\multirow{2}{*}{\textbf{GMAI}} & Shanghai AI Lab & \multirow{2}{*}{1,2,3} & \multicolumn{1}{c|}{TinyViT(\gls{S-TF})} & \multicolumn{1}{c|}{\multirow{2}{*}{Weighted Ensemble}} & \multirow{2}{*}{-} \\ \cline{4-4}
 & Shanghai, China & & \multicolumn{1}{c|}{EVA-02(\gls{S-TF})} & \multicolumn{1}{c|}{} &  \\ \Xhline{5\arrayrulewidth}
\multirow{3}{*}{\textbf{SANO}} & Sano Center for & \multirow{2}{*}{1,3} & \multicolumn{1}{c|}{\multirow{3}{*}{ResNet50(\gls{CNN})}} & \multicolumn{1}{c|}{\multirow{2}{*}{-}} & \multirow{2}{*}{-} \\
 & Computational Medicine & & \multicolumn{1}{c|}{} & \multicolumn{1}{c|}{} &  \\ \cline{3-3} \cline{5-6} 
 & Krakow, Poland & 2 & \multicolumn{1}{c|}{} & \multicolumn{1}{c|}{\gls{LSTM}\{5\}} & - \\ \Xhline{5\arrayrulewidth}
 & German Cancer & \multirow{3}{*}{2} & \multicolumn{1}{c|}{ResNet152(\gls{CNN})} & \multicolumn{1}{c|}{\gls{LSTM}\{15\}} & \multirow{3}{*}{Balanced Ensemble} \\ \cline{4-5}
\textbf{SDS-HD} & Research Center &  & \multicolumn{1}{c|}{EfficientNetB7(\gls{CNN})} & \multicolumn{1}{c|}{\gls{LSTM}\{15\}} &  \\ \cline{4-5}
\multicolumn{1}{|l|}{\textbf{}} & Heidelberg, Germany &  & \multicolumn{1}{c|}{SwinL\{1\}(\gls{S-TF})} & \multicolumn{1}{c|}{\gls{LSTM}\{12\}} &  \\ \Xhline{5\arrayrulewidth}
\multirow{3}{*}{\textbf{SK}} & Muroran Institute & 2 & \multicolumn{1}{c|}{\multirow{3}{*}{ConvNeXtTiny(\gls{CNN})}} & \multicolumn{1}{c|}{-} & - \\ \cline{3-3} \cline{5-6} 
 & of Technology & \multirow{2}{*}{3} & \multicolumn{1}{c|}{} & \multicolumn{1}{c|}{\multirow{2}{*}{\gls{LSTM}\{128\}$\langle$step$\rangle$}} & \multirow{2}{*}{-} \\
& Hokkaido, Japan &  & \multicolumn{1}{c|}{} & \multicolumn{1}{c|}{} &  \\ \Xhline{5\arrayrulewidth}
& National Center for & \multirow{3}{*}{1} & \multicolumn{1}{c|}{\multirow{3}{*}{ConvNeXtTiny(\gls{CNN})}} & \multicolumn{1}{c|}{\multirow{3}{*}{\gls{LSTM}\{512\}}} & \multirow{3}{*}{Threshold smoothing(\gls{TSF})} \\
\textbf{TSO-NCT} & Tumor Diseases &  & \multicolumn{1}{c|}{} & \multicolumn{1}{c|}{} &  \\ 
\textbf{} & Dresden, Germany &  & \multicolumn{1}{c|}{} & \multicolumn{1}{c|}{} &  \\ \Xhline{5\arrayrulewidth}
 & Universidad & \multirow{2}{*}{1} & \multicolumn{1}{c|}{MViT\{24\}(\gls{ST-TF})} & \multicolumn{1}{c|}{\multirow{2}{*}{StepFormer\{24$\times$8\}(ST-TF)}} & \multirow{2}{*}{Harmonic smoothing(\gls{TSF})} \\ \cline{4-4}
\textbf{UNI-} & de los Andes &  & \multicolumn{1}{c|}{DINO\{24\}(\gls{S-TF})} & \multicolumn{1}{c|}{} &  \\ \cline{3-6} 
\textbf{ANDES-23} & \multirow{2}{*}{Bogota, Colombia} & \multirow{2}{*}{2,3} & \multicolumn{1}{c|}{MViT\{24\}(\gls{ST-TF})} & \multicolumn{1}{c|}{FusionFormer} & Harmonic smoothing(\gls{TSF})$\langle$step$\rangle$ \\ \cline{4-4}
 & & & \multicolumn{1}{c|}{DINO\{24\}(\gls{S-TF})} & \multicolumn{1}{c|}{\{$24\times10\times2$\}(ST-TF)} & Threshold probability$\langle$instrument$\rangle$ \\ \hline
\end{tabular}
}
\caption{Team details (9-teams) and simplified model architectures for the successful 18-submissions. For the model columns, each row represents a different training component, and if a horizontal line is removed at a later stage it means the model features have been combined (e.g. in an Ensemble). () are given to indicate the type of model used for that stage. \{\} are given to indicate the window size of a temporal neural network (e.g. \{24\} represents 24-images have been turned into a sequence as an input). $\langle \rangle$ are given to indicate the task (step or instrument) for multi-task recognition if the same architecture is not used for both tasks. Citations: ARST \cite{Zou2022}; CSPDarknet53 \cite{Bochkovskiy2020}; ConvNeXtTiny \cite{Liu2022}; DenseNet201 \cite{Huang2016}, DINO \cite{Zhang2022}; EfficientNetB7 \cite{Tan2020}; EVA-02 \cite{Fang2024}; MViT \cite{Fan2021}; ResNet152, ResNet50 \cite{He2016}; Swin, SwinL \cite{Liu2021}; TeCNO \cite{Czempiel2020}, TinyViT \cite{Wu2022}, Threshold Smoothing \cite{Das2022}, XCiT \cite{Elnouby2021}.}
\label{tab:01teams}
\end{table*}

\begin{figure*}[t!]
    \centering
    \includegraphics[width=\textwidth]{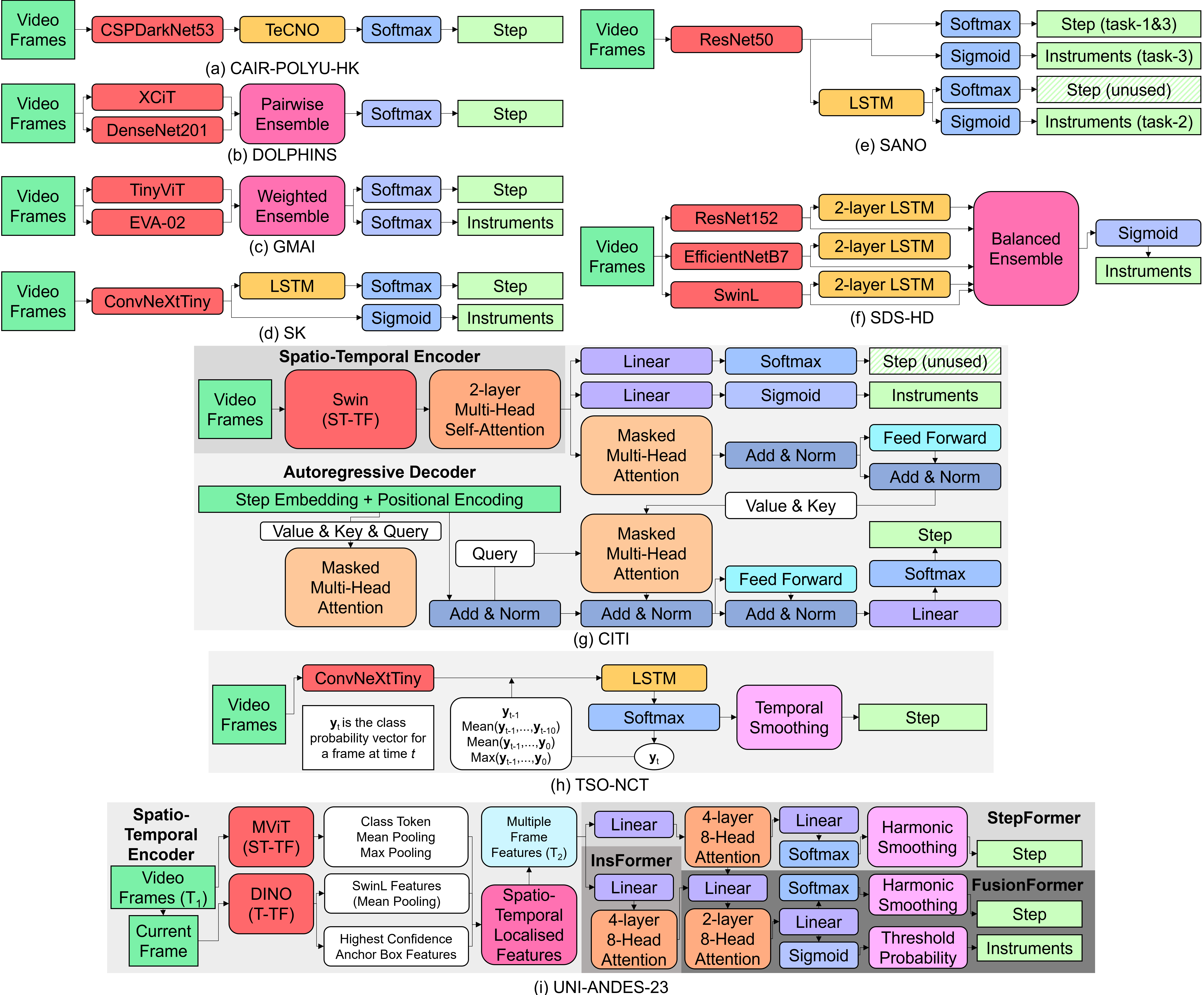}
    \caption{Architecture diagrams for all models. (a-f) represent models that use a spatial or spatial-temporal encoder followed by a temporal decoder, and (g-i) represent models that also utilise temporal propagation.}
    \label{fig:10models}
\end{figure*}

\begin{table*}[!t]
\centering
\resizebox{\textwidth}{!}{
\begin{tabular}{|c|c|ccc|c|ccc|ccc|c|cc|c|}
\hline
\multirow{2}{*}{\textbf{Team}} & \textbf{CAIR-} & \multicolumn{3}{c|}{\multirow{2}{*}{\textbf{CITI}}} & \multirow{2}{*}{\textbf{DOLPHINS}} & \multicolumn{3}{c|}{\multirow{2}{*}{\textbf{GMAI}}} & \multicolumn{3}{c|}{\multirow{2}{*}{\textbf{SANO}}} & \textbf{SDS-} & \multicolumn{2}{c|}{\multirow{2}{*}{\textbf{SK}}} & \textbf{TSO-} \\
 & \textbf{POLYU-HK} & \multicolumn{3}{c|}{} &  & \multicolumn{3}{c|}{} & \multicolumn{3}{c|}{} & \textbf{HD} & \multicolumn{2}{c|}{} & \textbf{NCT} \\ \hline
\textbf{Task} & \textbf{1} & \multicolumn{1}{c|}{\textbf{1\&3}} & \multicolumn{2}{c|}{\textbf{2\&3}} & \textbf{1} & \multicolumn{1}{c|}{\textbf{1}} & \multicolumn{1}{c|}{\textbf{2}} & \textbf{3} & \multicolumn{2}{c|}{\textbf{1\&3}} & \textbf{2} & \textbf{2} & \multicolumn{1}{c|}{\textbf{2}} & \textbf{3} & \textbf{1} \\ \hline
\textbf{Loss} & CE & \multicolumn{3}{c|}{CE} & CE & \multicolumn{3}{c|}{CE} & \multicolumn{2}{c|}{CE$/$BCE} & CE$/$BCE & BCE & \multicolumn{2}{c|}{CE} & CE$|$TS \\ \hline
\textbf{Activation} & ReLU & \multicolumn{3}{c|}{ReLU} & ReLU & \multicolumn{3}{c|}{ReLU} & \multicolumn{2}{c|}{ReLU} & Softmax & ReLU & \multicolumn{2}{c|}{GeLU} & GeLU$|$Sigmoid \\ \hline
\textbf{Final activation} & Softmax & \multicolumn{1}{c|}{Softmax} & \multicolumn{2}{c|}{Sigmoid}  & Softmax & \multicolumn{3}{c|}{Softmax} & \multicolumn{2}{c|}{Softmax} & Sigmoid & Sigmoid & \multicolumn{1}{c|}{Sigmoid} & \multicolumn{1}{c|}{Softmax} & Softmax \\ \hline
\textbf{Pre-trained} & ImageNet & \multicolumn{3}{c|}{-} & ImageNet & \multicolumn{3}{c|}{ImageNet} & \multicolumn{3}{c|}{ImageNet} & ImageNet & \multicolumn{2}{c|}{ImageNet} & ImageNet \\ \hline
\textbf{Multitask training} & - & \multicolumn{3}{c|}{Yes} & - & \multicolumn{3}{c|}{Yes} & \multicolumn{3}{c|}{Yes} & - & \multicolumn{1}{c|}{-} & Yes & - \\ \hline
\textbf{Temporal training} & ETE & \multicolumn{1}{c|}{Sep} & \multicolumn{2}{c|}{ETE} & - & \multicolumn{3}{c|}{-} & \multicolumn{2}{c|}{-} & \multicolumn{1}{c|}{Sep} & \multicolumn{1}{c|}{Sep} & \multicolumn{1}{c|}{-} & Sep & ETE \\ \hline
\textbf{Removed borders} & Yes & \multicolumn{3}{c|}{Yes} & - & \multicolumn{3}{c|}{-} & \multicolumn{3}{c|}{Yes} & Yes & \multicolumn{2}{c|}{-} & - \\ \hline
\textbf{Augmentation} & \multirow{2}{*}{1.0} & \multicolumn{3}{c|}{\multirow{2}{*}{1.0}} & \multirow{2}{*}{1.0} & \multicolumn{3}{c|}{\multirow{2}{*}{1.0}} & \multicolumn{3}{c|}{\multirow{2}{*}{1.0}} & \multirow{2}{*}{0.5} & \multicolumn{2}{c|}{\multirow{2}{*}{1.0}} & \multirow{2}{*}{0.5} \\
\textbf{probability} &  & \multicolumn{3}{c|}{} &  & \multicolumn{3}{c|}{} & \multicolumn{3}{c|}{} &  & \multicolumn{2}{c|}{} &  \\ \hline
\textbf{Resizing (pixels)} & $256\times448$ & \multicolumn{3}{c|}{$192\times192$} & $224\times224$ & \multicolumn{3}{c|}{$224\times224$} & \multicolumn{3}{c|}{$224\times224$} & $384\times384$ & \multicolumn{2}{c|}{$224\times224$} & $216\times384$ \\ \hline
\textbf{Rotation (degrees)} & - & \multicolumn{3}{c|}{-} & - & \multicolumn{3}{c|}{-} & \multicolumn{3}{c|}{-} & $\pm45$ & \multicolumn{2}{c|}{$\pm5$} & $\pm15$ \\ \hline
\textbf{Reflection} & - & \multicolumn{3}{c|}{Horizontal} & Horizontal & \multicolumn{3}{c|}{Horizontal} & \multicolumn{3}{c|}{-} & Horizontal\&Vertical & \multicolumn{2}{c|}{-} & - \\ \hline
\textbf{Translation (x\&y)} & - & \multicolumn{3}{c|}{-} & - & \multicolumn{3}{c|}{-} & \multicolumn{3}{c|}{-} & - & \multicolumn{2}{c|}{$\pm5\%$} & $\pm5\%$ \\ \hline
\textbf{Scaling} & - & \multicolumn{3}{c|}{-} & - & \multicolumn{3}{c|}{-} & \multicolumn{3}{c|}{-} & $\pm10\%$ & \multicolumn{2}{c|}{$\pm5\%$} & $\pm5\%$ \\ \hline
\multirow{3}{*}{\textbf{Colour}} & \multirow{3}{*}{-} & \multicolumn{3}{c|}{ImageNet} & ImageNet & \multicolumn{3}{c|}{\multirow{3}{*}{-}} & \multicolumn{3}{c|}{ImageNet} & Colour jitter & \multicolumn{2}{c|}{Blur} & RBG$\pm15$ \\ \cline{13-16} 
 &  & \multicolumn{3}{c|}{Normal-} & Normal- & \multicolumn{3}{c|}{} & \multicolumn{3}{c|}{Normal-} & Contrast & \multicolumn{2}{c|}{HSV} & Contrast \\
 &  & \multicolumn{3}{c|}{isation} & isation & \multicolumn{3}{c|}{} & \multicolumn{3}{c|}{isation} & equalisation & \multicolumn{2}{c|}{augmentations} & $\pm0.2$ \\ \hline
\textbf{Data balancing} & - & \multicolumn{3}{c|}{-} & - & \multicolumn{3}{c|}{-} & \multicolumn{3}{c|}{-} & Instrument upsampling & \multicolumn{2}{c|}{-} & - \\ \hline
\textbf{Validation} & Suggested & \multicolumn{3}{c|}{Suggested} & Suggested & \multicolumn{3}{c|}{-} & \multicolumn{3}{c|}{Suggested} & 5-fold & \multicolumn{2}{c|}{12,15,17,20,22} & Suggested \\ \hline
\textbf{Training shuffling} & Yes & \multicolumn{3}{c|}{Yes} & Yes & \multicolumn{3}{c|}{Yes} & \multicolumn{3}{c|}{Yes} & Yes & \multicolumn{2}{c|}{Yes} & No \\ \hline
\textbf{Val shuffling} & No & \multicolumn{3}{c|}{No} & No & \multicolumn{3}{c|}{No} & \multicolumn{3}{c|}{Yes} & No & \multicolumn{2}{c|}{No} & No \\ \hline
\textbf{Trained epochs} & 30 & \multicolumn{1}{c|}{10} & \multicolumn{2}{c|}{8} & 50 & \multicolumn{3}{c|}{20} & \multicolumn{3}{c|}{40} & 10 & \multicolumn{2}{c|}{50} & 200 \\ \hline
\textbf{Evaluation metric} & Task & \multicolumn{3}{c|}{Task} & Task & \multicolumn{3}{c|}{-} & \multicolumn{2}{c|}{\F{}} & Task & \F{}+mAP & \multicolumn{2}{c|}{Minimal loss} & \F{} \\ \hline
\textbf{Best model choice} & Val & \multicolumn{3}{c|}{Val} & Val & \multicolumn{3}{c|}{Last epoch} & \multicolumn{3}{c|}{Val} & Val & \multicolumn{2}{c|}{Val} & Val \\ \hline
\textbf{Batch size} & 200 & \multicolumn{1}{c|}{Video} & \multicolumn{2}{c|}{4} & 25 & \multicolumn{3}{c|}{16} & \multicolumn{3}{c|}{128} & 64 & \multicolumn{1}{c|}{128} & 32 & 512 \\ \hline
\textbf{Training hours} & 40 & \multicolumn{1}{c|}{4} & \multicolumn{2}{c|}{24} & 12 & \multicolumn{3}{c|}{10} & \multicolumn{3}{c|}{2} & 88 & \multicolumn{1}{c|}{3} & 64 & 48 \\ \hline
\textbf{Backpropogation} & SGD & \multicolumn{3}{c|}{Adam} & Adam & \multicolumn{3}{c|}{AdamW} & \multicolumn{3}{c|}{SGD} & Adam & \multicolumn{2}{c|}{Adam} & AdamW \\ \hline
\textbf{Learning} & \multirow{2}{*}{1E-3} & \multicolumn{3}{c|}{\multirow{2}{*}{1E-4}} & \multirow{2}{*}{1E-3} & \multicolumn{3}{c|}{\multirow{2}{*}{1E-3}} & \multicolumn{2}{c|}{\multirow{2}{*}{1E-3}} & \multicolumn{1}{c|}{\multirow{2}{*}{5E-3}} & 2E-4 & \multicolumn{1}{c|}{\multirow{2}{*}{1E-4}} & \multirow{2}{*}{1E-5} & \multirow{2}{*}{5E-4} \\
\textbf{rate} &  & \multicolumn{3}{c|}{} &  & \multicolumn{3}{c|}{} & \multicolumn{2}{c|}{} & & ($>$2E-5) & \multicolumn{1}{c|}{} &  &  \\ \hline
\textbf{Momentum} & 9E-2 & \multicolumn{3}{c|}{-} & - & \multicolumn{3}{c|}{-} & \multicolumn{3}{c|}{9E-2} & - & \multicolumn{2}{c|}{-} & - \\ \hline
\textbf{Decay} & - & \multicolumn{3}{c|}{1E-3} & - & \multicolumn{3}{c|}{-} & \multicolumn{3}{c|}{-} & 1E-6 & \multicolumn{2}{c|}{-} & 1E-2 \\ \hline
\textbf{GPU (NVIDIA)} & A100 & \multicolumn{3}{c|}{TITAN RTX} & RTX A6000 & \multicolumn{3}{c|}{V100} & \multicolumn{3}{c|}{A100} & V100 & \multicolumn{2}{c|}{RTX4090} & RTX A5000 \\ \hline
\textbf{GPU (GB)} & 80 & \multicolumn{3}{c|}{24} & 48 & \multicolumn{3}{c|}{32} & \multicolumn{3}{c|}{$2\times80$} & 32 & \multicolumn{2}{c|}{24} & 24 \\ \hline
\end{tabular}
}
\caption{Training parameters and augmentations utilised by the models excluding UNI-ANDES-23. \Q{$/$} implies implementation details for steps or instruments (e.g. CE/BCE means CE used for steps and BCE used for instruments). \Q{$|$} implies implementation details from stage-1 to stage-2 (e.g. GeLU$|$Sigmoid means GeLU used for stage-1 and Sigmoid used for stage-2). Abbreviations: Adam (Adaptive Moment Estimation), BCE (Binary Cross-Entropy Loss Function), CE (Cross-Entropy Loss Function), ETE (End To End Temporal Training), GeLU (Gaussian error Linear Unit), HSV (Hue Saturation Value), mAP (mean Average Precision), RBG (Red Blue Green), ReLU (Rectified Linear Unit), SGD (Stochastic Gradient Descent), TS (Temporal Smoothing Loss Function), Sep (Separate Temporal Training), Val (Validation Dataset).}
\label{tab:03train}
\end{table*}

\tab{tab:01teams} displays a summary of the 9-teams from 6-countries, and the corresponding 18-submissions: 7 for Task-1; 6 for Task-2; and 5 for Task-3. All models use either a \gls{S-E} (\gls{CNN}; \gls{S-TF}) or \gls{ST-E}  (\gls{ST-TF}), with the majority using a temporal decoder (\gls{LSTM}; \gls{TCN}; \gls{T-TF}), and a few perform online post-processing (\gls{TSF}). There are some which use multiple neural networks and combine them via an Ensemble. Architectural diagrams of all models are displayed in \fig{fig:10models}.

\mbox{Tables \ref{tab:03train} and \ref{tab:04uni}} display a summary of the training parameters and image augmentations. Although there are a few commonalities between the methods (\gls{CE} loss function; resizing input images), there are vast differences. The majority do not implement strong image augmentations; or any data balancing, whereas a majority do use the suggested validation split; pre-train on ImageNet; or use \gls{Adam} for backpropogation. The remaining parameters are even: some use \gls{ReLU}; some remove the black borders of an image; and some use the task evaluation metric.

Below is an overview of each model:

\subsection{CAIR-POLYU-HK}
CAIR-POLYU-HK consisted of You Pang; Zhen Chen; Xiaobo Qiu; and Zhen Sun, from the Hong Kong Institute of Science and Innovation, China.

For task-1, their model consisted of 2-stages: a cross stage partial \gls{CNN} (CSPDarknet53 \cite{Bochkovskiy2020}); followed by a 2-layer 10-window \gls{TCN} (TeCNO \cite{Czempiel2020}).

CAIR-POLYU-HK had the largest batch size of 200, utilising an 80-GB NVIDIA-A100.

\subsection{CITI}
CITI consisted of Xiaoyang Zou; and Guoyan Zheng, from Shanghai Jiao Tong University, China.

For the 3-tasks a \gls{ST-E}; plus autoregressive decoder (ARST \cite{Zou2022}) was used. The \gls{ST-E} took a 20-window sequential video frame input, outputting both step (just for training) and instrument (task-2\&3) classifications. It comprised of a \gls{ST-TF} (Swin \cite{Liu2021}) followed by a 2-layer \gls{MHSA} \cite{Zou2024}.

ARST took an 80-window input comprising of frame-wise visual features extracted by \gls{ST-E} and shifted step outputs, outputting step classifications. It comprised of an initial \gls{MMHA}, followed by a mutual \gls{MMHA} taking the Value and Key output of the \gls{ST-E} after it has passed through a \gls{MMHA} and the Query output from the initial \gls{MMHA} (also passed to the normalisation layer). Positional encoding is added to embed the frame position for each step (as defined in \cite{Vaswani2017}). 

\tab{tab:03train} CITI task-2\&3 and task-1\&3  represent the \gls{ST-E} and ARST training parameters respectively.

\subsection{DOLPHINS}
DOLPHINS consisted of Abdul Qayyum; Moona Mazher; Imran Razzak; and Steven Niederer, from Imperial College London, United Kingdom.

For task-1, their model consisted of 2-stages: a cross variance \gls{S-TF} (XCiT \cite{Elnouby2021}) and a \gls{CNN} (DenseNet201 \cite{Huang2016}); fused via pairwise ensemble.

\subsection{GMAI}
GMAI consisted of Tianbin Li; Jin Ye; Junjun He; Yanzhou Su; Pengcheng Chen; and Junlong Cheng, from the Shanghai Artificial Intelligence Lab, China.

For all 3-tasks, their model consisted of 2-stages: a \gls{S-TF} utilising fast knowledge distillation (TinyViT \cite{Wu2022}) and another \gls{S-TF} utilising masked image modeling (EVA-02 \cite{Fang2024}); fused via weighted ensemble.

\newpage 

\subsection{SANO}
SANO consisted of Szymon Płotka; and Joanna Kaleta, from the Sano Center for Computational Medicine, Poland.

For tasks-1\&3 their model consisted of 1-stage: a residual \gls{CNN} (ResNet50 \cite{He2016}) for step (task-1\&3) and instrument (task-3) classification. 

For task-2 their model consisted of 2-stages: the trained \gls{CNN} was frozen; followed by a 5-window \gls{LSTM} for both instrument (task-2) and step (just for training) classification. The details in \tab{tab:03train} SANO task-2 represent the \gls{LSTM} training parameters.

\subsection{SDS-HD} 
SDS-HD consisted of Amine Yamlahi; Antoine Jund; Finn-Henri Smidt; Patrick Godau; and Lena Maier-Hein, from the German Cancer Research Center, Germany.

For task-2, their model consisted of 3-stages: 3-encoders (ResNet152 \cite{He2016}, EfficientNetB7 \cite{Tan2020}, SwinL \cite{Liu2021}); with their respective spatial features each fed into separate 2-layer \glspl{LSTM} with 0.2-dropout (15-window, 15-window, 12-window); the outputs of which were fused together via a balanced ensemble, consisting of the encoders' and \glspl{LSTM}' predictions.

SDS-HD used a variety of alternative training techniques when compared to the other participants. Firstly, they balanced the data: 5-instrument classes (07; 10; 11; 12; 15) were upsampled and the remaining classes were downsampled. Secondly, they introduced both horizontal and vertical reflections, along with colour augmentations: colour jitter by modifying hue; saturation; and brightness, in addition to \gls{CLAHE} augmentation. Thirdly, they utilised \gls{mAP} as an alternative evaluation metric along with the task specific macro-\F{}. Finally, Adam backpropogation was enhanced via cosine annealing with a learning rate of 2E-4, with a minimum of 2E-5 and a 1E-6 decay rate.

\subsection{SK}
SK consisted of Satoshi Kondo; Satoshi Kasai; and Kousuke Hirasawa, from Muroran Institute of Technology, Niigata University of Health and Welfare, and Konica Minolta, Inc., Japan, respectively.

For task-2, their model consisted of 1-stage: a \gls{CNN} (ConvNeXtTiny \cite{Liu2022}) for instrument classification. For task-3, their model consisted of 2-stages: the trained \gls{CNN} was frozen for instrument classification; and a 128-window \gls{LSTM} was added for step classification. The details in \tab{tab:03train} SK task-3 represent the \gls{LSTM} training parameters.

\newpage

\subsection{TSO-NCT}
TSO-NCT consisted of Dominik Rivoir, from the National Center for Tumor Diseases, Germany.

For task-1, their model  consisted of 3-stages: a \gls{CNN}  (ConvNeXtTiny \cite{Liu2022}); a 512-window \gls{LSTM}; and a 7-window \gls{TSF} (Threshold Smoothing \cite{Das2022}).

Inspired by \gls{SSM} \cite{Ban2021}, to propagate temporal features, for each frame, the softmax class scores of: the previous frame; the mean of the previous 10-frames, the mean and maximum of all previous frames, were fed into the \gls{LSTM} in addition to the \gls{CNN} spatial features. Per video, all temporal features (softmax scores and \gls{LSTM} hidden state) are propagated across the unshuffled batches.

Threshold smoothing ensures a class transition only takes place after it has been predicted for a sufficient number of frames (in this case 7), otherwise it is left unchanged. In doing so, prediction consistency is improved in aims to increase \E{}. Any steps not considered for evaluation (i.e. steps -1; 11; 13) were replaced with the most recent permitted step.

\subsection{UNI-ANDES-23}
UNI-ANDES-23 consisted of Alejandra Pérez; Santiago Rodriguez; Pablo Arbeláez; Nicolás Ayobi; and Nicolás Aparicio from Universidad de los Andes, Colombia.

For all 3-tasks, their model consisted of 3-stages: a \gls{ST-E}; a \gls{ST-D}; and Harmonic Smoothing or Threshold Probability for step or instrument classification respectively.

In stage-1 for all 3-tasks, the \gls{ST-E} is composed of two concatenated transformers. The first is a 24-window (6-seconds$\times$4-\gls{fps}) \gls{ST-TF} (MViT \cite{Fan2021}), concatenating the class token; mean pooled features; and max pooled features. The second is a \gls{S-TF} (DINO \cite{Zhang2022}) acting on the final frame using SwinL \cite{Liu2021}, concatenating global max pooled features; and localised instrument features via anchor boxes.

For task-1, the \gls{ST-D} (StepFormer) consists of an 8-window 4-layer 8-head attention transformer. For task-2, the \gls{ST-D} (FusionFormer) consists of an identical transformer (InsFormer) combined with StepFormer (frozen weights) via a 2-layer 8-head attention transformer. For task-3, both StepFormer and InsFormer have frozen weights.

Harmonic Smoothing is an online post-processing \gls{TSF} defined as follows: given the class probability vector of the current ($\textbf{y}_\text{t}$) and previous frame ($\textbf{y}_\text{t-1}$), if  $\mathrm{max}\{\textbf{y}_\text{t}\}<\mathrm{max}\{\textbf{y}_\text{t-1}\}$, then $\hat{\textbf{y}}_\text{t}=2\left(\textbf{y}_\text{t}^{-1} + \textbf{y}_\text{t-1}^{-1}\right)^{-1}$ where $\hat{\textbf{y}}_\text{t}$ is the updated class probability vector. This function is repeated for 750-iterations for improved temporal consistency, before the usual argmax function is applied for a final classification. Any steps not considered for evaluation were removed at this stage.

Threshold Probability is an online post-processing function defined as follows: if the second highest value in the class probability vector is less than 0.4, then only predict the first highest value's corresponding class; if at least two of the highest values in this vector are greater than or equal to 0.4 and this includes the value corresponding to the background class, then predict the two highest values' classes excluding the background class; in all other cases predict the two highest values' corresponding classes.

\begin{table}[!t]
\centering
\resizebox{\columnwidth}{!}{
\begin{tabular}{|c|ccccc|}
\hline
\textbf{Network} & \multicolumn{1}{c|}{\textbf{MViT}} & \multicolumn{1}{c|}{\textbf{DINO}} & \multicolumn{1}{c|}{\textbf{StepFormer}} & \multicolumn{1}{c|}{\textbf{InsFormer}} & \textbf{FusionFormer} \\ \hline
\textbf{Loss} & \multicolumn{1}{c|}{CE} & \multicolumn{1}{c|}{CE} & \multicolumn{1}{c|}{CE} & \multicolumn{1}{c|}{BCE} & BCE \\ \hline
\textbf{Activation} & \multicolumn{1}{c|}{ReLU} & \multicolumn{1}{c|}{ReLU} & \multicolumn{1}{c|}{GeLU} & \multicolumn{1}{c|}{GeLU} & GeLU \\ \hline
\textbf{Final activation} & \multicolumn{1}{c|}{-} & \multicolumn{1}{c|}{-} & \multicolumn{1}{c|}{Softmax} & \multicolumn{1}{c|}{Sigmoid} & Softmax/Sigmoid \\ \hline
\multirow{2}{*}{\textbf{Pre-trained}} & \multicolumn{1}{c|}{Kinetics400} & \multicolumn{1}{c|}{\multirow{2}{*}{COCO}} & \multicolumn{1}{c|}{\multirow{2}{*}{-}} & \multicolumn{1}{c|}{\multirow{2}{*}{-}} & \multirow{2}{*}{-} \\
 & \multicolumn{1}{c|}{+ PSI-AVA} & \multicolumn{1}{c|}{} & \multicolumn{1}{c|}{} & \multicolumn{1}{c|}{} &  \\ \hline
\textbf{Temporal training} & \multicolumn{5}{c|}{Yes} \\ \hline
\textbf{Multitask training} & \multicolumn{5}{c|}{Yes} \\ \hline
\textbf{Removed borders} & \multicolumn{2}{c|}{Yes} & \multicolumn{3}{c|}{-} \\ \hline
\textbf{Augmentation} & \multicolumn{1}{c|}{\multirow{2}{*}{1.0}} & \multicolumn{1}{c|}{\multirow{2}{*}{1.0}} & \multicolumn{3}{c|}{\multirow{2}{*}{-}} \\ \cline{1-1}
\textbf{probability} & \multicolumn{1}{c|}{} & \multicolumn{1}{c|}{} & \multicolumn{3}{c|}{} \\ \hline
\textbf{Resizing (pixels)} & \multicolumn{1}{c|}{$224\times224$} & \multicolumn{1}{c|}{$894\times800$} & \multicolumn{3}{c|}{$805\times720$} \\ \hline
\textbf{Rotation (degrees)} & \multicolumn{5}{c|}{-} \\ \hline
\textbf{Reflection} & \multicolumn{5}{c|}{-} \\ \hline
\textbf{Translation (x\&y)} & \multicolumn{1}{c|}{-} & \multicolumn{1}{c|}{Yes} & \multicolumn{3}{c|}{-} \\ \hline
\textbf{Scaling} & \multicolumn{1}{c|}{-} & \multicolumn{1}{c|}{Yes} & \multicolumn{3}{c|}{-} \\ \hline
\textbf{Colour} & \multicolumn{1}{c|}{Jitter (0.4)} & \multicolumn{1}{c|}{-} & \multicolumn{3}{c|}{-} \\ \hline
\textbf{Data} & \multicolumn{1}{c|}{Weighted} & \multicolumn{2}{c|}{Weights inverse} & \multicolumn{2}{c|}{Weighted loss} \\
\textbf{balancing} & \multicolumn{1}{c|}{sampling} & \multicolumn{2}{c|}{of sample size} & \multicolumn{2}{c|}{2$\times$(step1,step14)} \\ \hline
\textbf{Validation} & \multicolumn{1}{c|}{} & \multicolumn{1}{c|}{} & \multicolumn{1}{c|}{} & \multicolumn{1}{c|}{} &  \\ \hline
\textbf{Training shuffling} & \multicolumn{5}{c|}{No} \\ \hline
\textbf{Val shuffling} & \multicolumn{5}{c|}{No} \\ \hline
\textbf{Trained epochs} & \multicolumn{1}{c|}{16} & \multicolumn{1}{c|}{12} & \multicolumn{3}{c|}{50} \\ \hline
\textbf{Evaluation metric} & \multicolumn{5}{c|}{Task} \\ \hline
\textbf{Best model choice} & \multicolumn{5}{c|}{Val} \\ \hline
\textbf{Batch size} & \multicolumn{1}{c|}{12} & \multicolumn{1}{c|}{4} & \multicolumn{3}{c|}{3000} \\ \hline
\textbf{Training hours} & \multicolumn{1}{c|}{64} & \multicolumn{1}{c|}{12} & \multicolumn{3}{c|}{8} \\ \hline
\textbf{Backpropogation} & \multicolumn{1}{c|}{SGD} & \multicolumn{1}{c|}{AdamW} & \multicolumn{1}{c|}{Adam} & \multicolumn{1}{c|}{Lion} & Adam \\ \hline
\textbf{Learning} & \multicolumn{1}{c|}{\multirow{2}{*}{1.25E-2}} & \multicolumn{1}{c|}{\multirow{2}{*}{1E-4}} & \multicolumn{1}{c|}{\multirow{2}{*}{1E-4}} & \multicolumn{1}{c|}{1E-5} & \multirow{2}{*}{1E-4} \\
\textbf{rate} & \multicolumn{1}{c|}{} & \multicolumn{1}{c|}{} & \multicolumn{1}{c|}{} & \multicolumn{1}{c|}{(Adam 1E-4)} &  \\ \hline
\textbf{Momentum} & \multicolumn{1}{c|}{0.9} & \multicolumn{1}{c|}{-} & \multicolumn{1}{c|}{-} & \multicolumn{1}{c|}{-} & - \\ \hline
\textbf{Decay} & \multicolumn{1}{c|}{-} & \multicolumn{1}{c|}{1E-4} & \multicolumn{1}{c|}{-} & \multicolumn{1}{c|}{1E-2} & - \\ \hline
\textbf{GPU (NVIDIA)} & \multicolumn{5}{c|}{Quadro RTX8000} \\ \hline
\textbf{GPU (GB)} & \multicolumn{5}{c|}{48GB} \\ \hline
\end{tabular}
}
\caption{Training parameters and augmentations utilised by UNI-ANDES-23.}
\label{tab:04uni}
\end{table}

\begin{figure*}[!t]
    \centering
    \begin{subfigure}[b]{0.49\textwidth}
        \centering
        \includegraphics[width=\columnwidth]{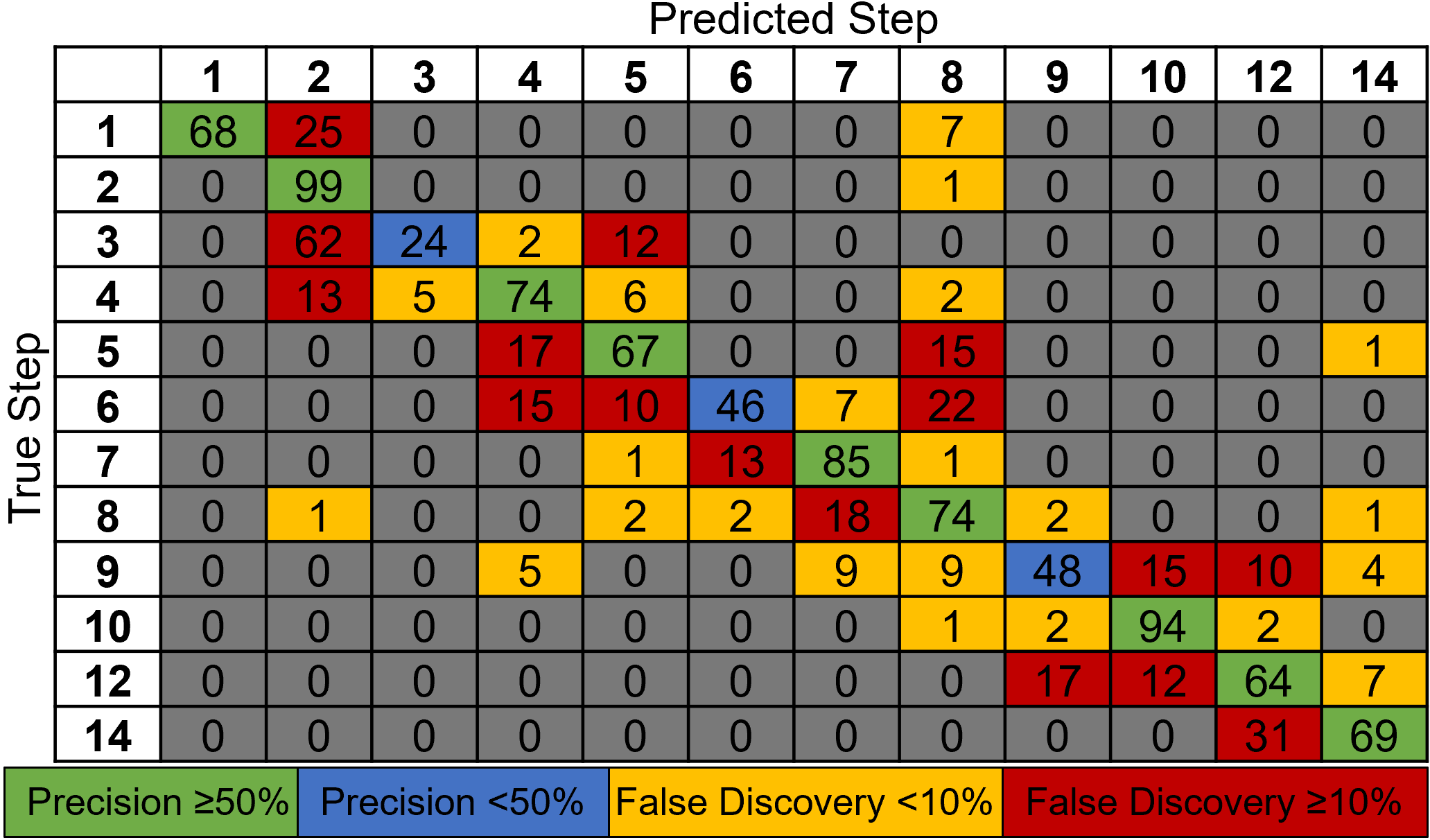}
        \caption{CITI's task-1 (1\textsuperscript{st}) and task-3 (1\textsuperscript{st}) model.}
        \label{fig:11asteps_confusion_citi}
    \end{subfigure}
    \hfill
    \begin{subfigure}[b]{0.49\textwidth}
        \centering
        \includegraphics[width=\columnwidth]{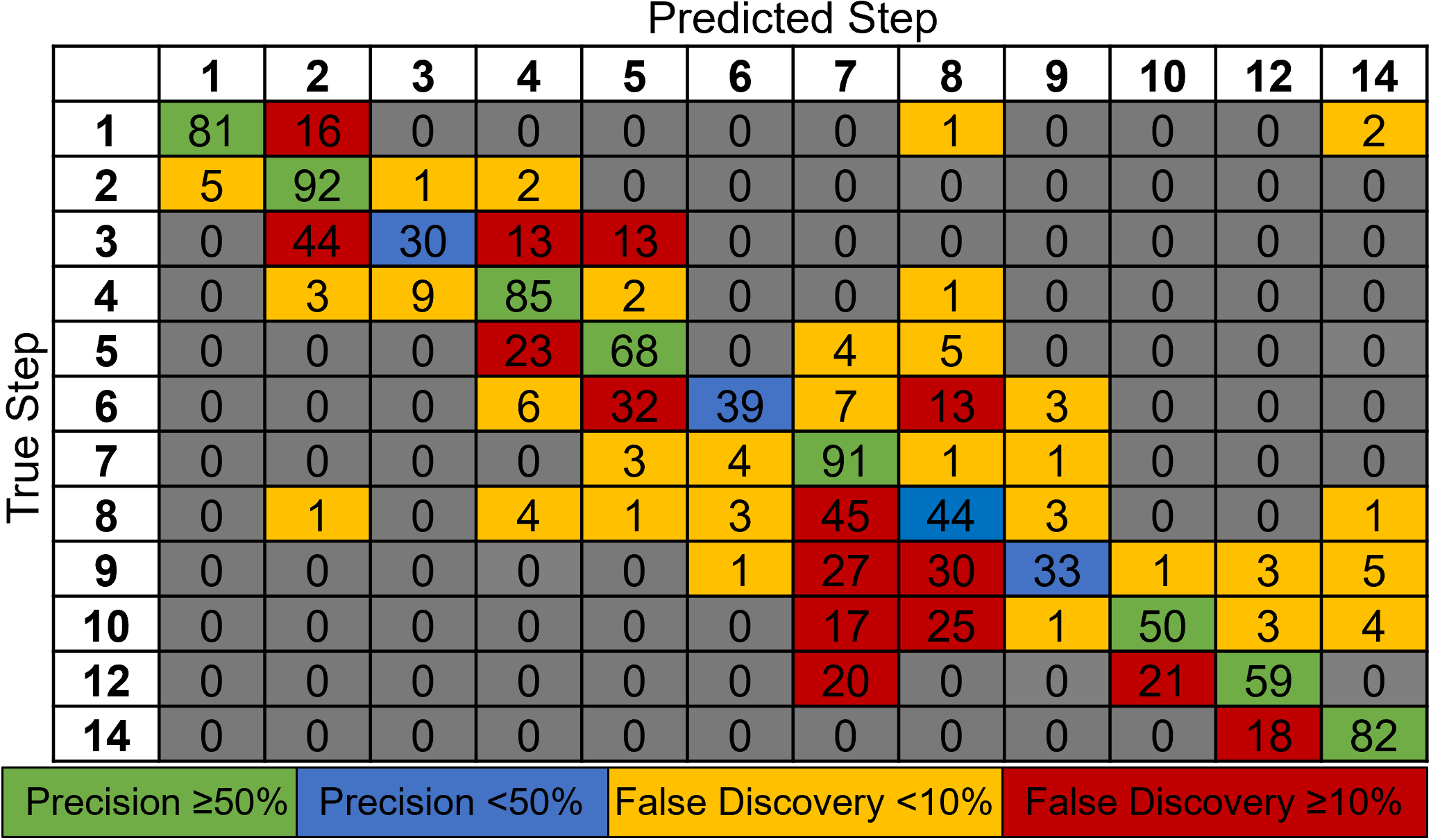}
        \caption{TSO-NCT's task-1 (2\textsuperscript{nd}) model.}
        \label{fig:11bsteps_confusion_tso}
    \end{subfigure}
    \begin{subfigure}[b]{0.49\textwidth}
        \centering
        \includegraphics[width=\columnwidth]{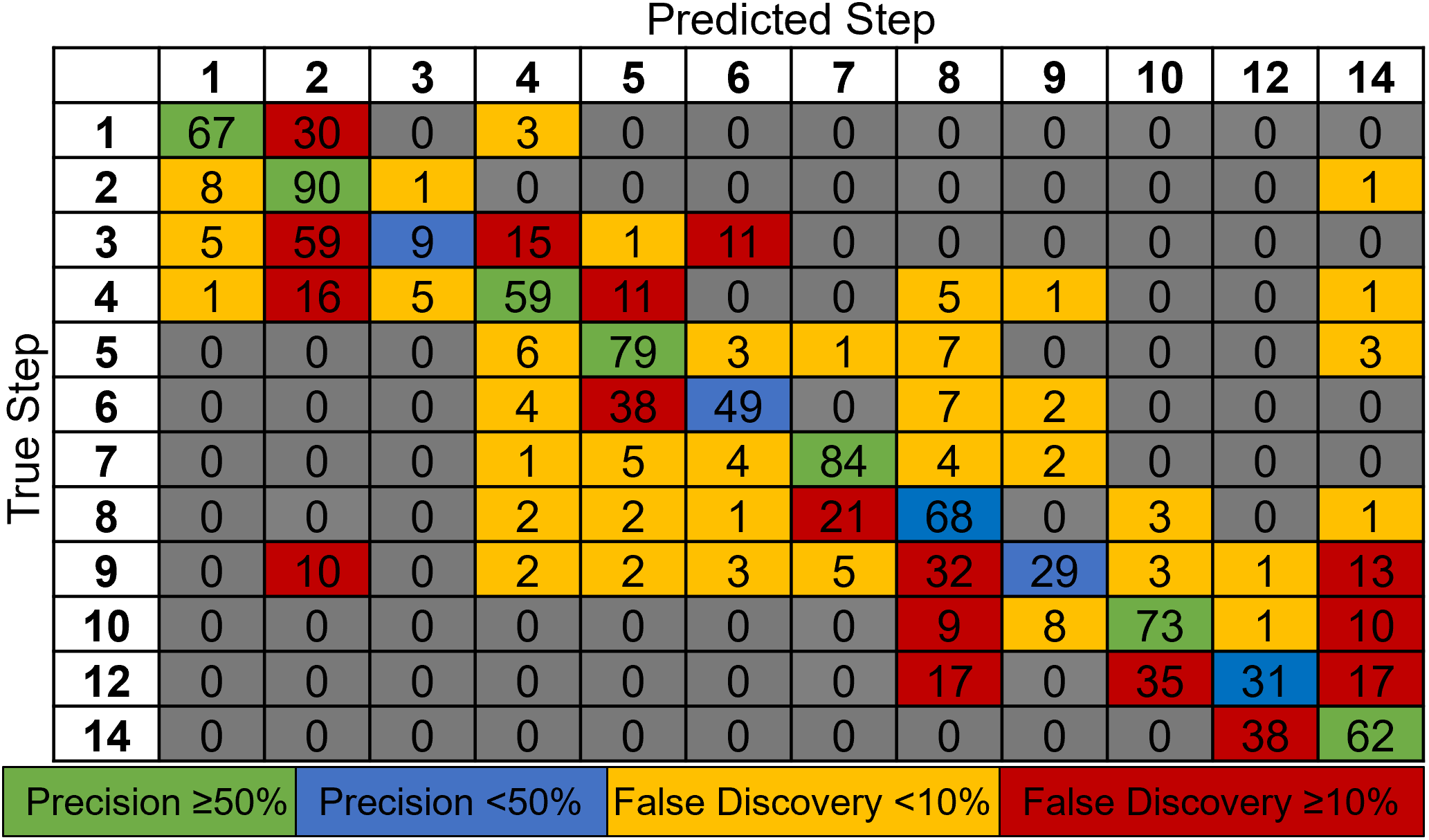}
        \caption{UNI-ANDES-23's task-3 (2\textsuperscript{nd}) model.}
        \label{fig:11csteps_confusion_uni}
    \end{subfigure}
    \begin{subfigure}[b]{0.49\textwidth}
        \centering
        \includegraphics[width=\columnwidth]{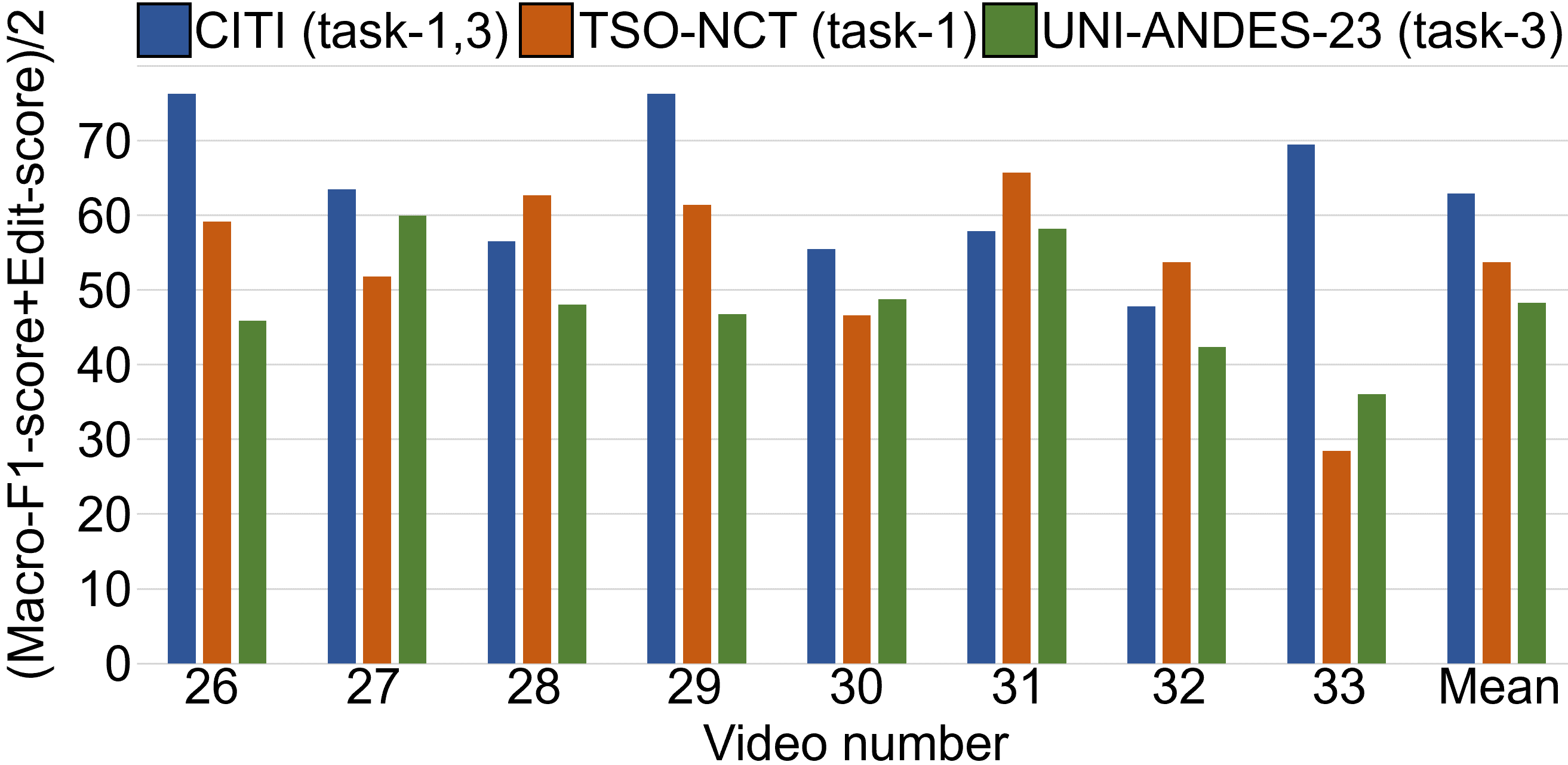}
        \caption{Step recognition results for each testing video.}
        \label{fig:11dsteps_test}
    \end{subfigure}
    \caption{In-depth details of the top models in step recognition:    
    (a-c) Confusion matrices, mean-averaged across the 8-testing-videos. (d) Per-video performance.}
    \label{fig:task1}
\end{figure*}

\section{Results \& Discussion} \label{sec:results}
\subsection{Ranking method}
Each video is considered one case of equal value, hence the rankings are determined by the tasks' evaluation metric mean-averaged across the 8-testing-videos (no missing results).

\subsection{Task-1}

\begin{table}[!t]
\centering
\resizebox{\columnwidth}{!}{
\begin{tabular}{c|c|c|c|c}
 & \multirow{2}{*}{Team} & (Macro-\F{} & Macro- & Edit- \\ 
 &  & + \E{})/2 & \F{} & score \\ \hline
1 & CITI & 62.9±09.7 & 61.1±10.6 & 64.7±10.1 \\
2 & TSO-NCT & 53.7±11.2 & 58.2±10.9 & 49.2±13.0 \\
3 & UNI-ANDES-23 & 48.3±07.3 & 50.1±09.3 & 46.5±08.2 \\
4 & SANO & 20.5±03.2 & 39.6±06.5 & 01.4±00.4 \\
5 & DOLPHINS & 15.2±04.0 & 28.9±08.2 & 01.6±00.7 \\
6 & GMAI & 03.7±00.2 & 06.8±00.3 & 00.5±00.1 \\
7 & CAIR-POLYU-HK & 03.5±00.8 & 05.8±01.5 & 01.1±00.3 \\
\end{tabular}
}
\caption{12-steps multi-class online recognition (task-1) rankings. Metrics are calculated across the 8-testing-videos (mean±std).}
\label{tab:05task1}
\end{table}

Results for the 7-submissions to 12-steps multi-class online recognition are displayed in \tab{tab:05task1}, with £700 and £300 awarded to 1\textsuperscript{st} and 2\textsuperscript{nd} places respectively.

There is a strong performance, with the best models achieving 63\% (CITI) and 54\% (TSO-NCT) on the task metric. Macro-\F{} is high, with the top 3-models achieving $>50\%$, although there is a slow decline with the bottom 2-models achieving $<7\%$. There is large variance in \E{}, with the top 3-models achieving $>46\%$, and the remaining $<2\%$.

Although the best models use different architectures, a commonality between them is the use of propagating temporal features. For CITI and UNI-ANDES-23 via positional encoding, and for TSO-NCT via feeding classification vectors of previous frames back into the \gls{LSTM} hidden state. It is clear models with temporal decoders and \glspl{TSF} outperform those that are purely spatial, both in frame-level classification and significantly in temporal consistency.

For the top models \gls{std} is $\approx10\%$, as can be more clearly seen in \fig{fig:11dsteps_test}. Although there is some variance between videos, they performance is generally similar. In videos 26; 29; 33 CITI significantly outperforms the other models, whereas TSO-NCT outperforms CITI in videos 28; 31; 32. The differences between the models, as well as between videos, highlights the difficulty of creating a generalised model.

\fig{fig:11asteps_confusion_citi} and \fig{fig:11bsteps_confusion_tso} displays the step confusion matrix for CITI and TSO-NCT repsectively. Steps are often predicted as a neighbouring step, which is expected (\fig{fig:08steps_transition}). Step-8 (haemostasis) is special as it is used sporadically for short periods during a surgery, and therefore other steps are often predicted as it. The biggest difference between the models is overpredicting the dominant class step-7 (tumour excision) in TSO-NCT. Across both models there is poor performance for steps 3; 6; 9, suggesting these are inherently difficult steps to classify.

\subsection{Task-2}
Results for the 6-submissions to 19-instruments multi-label online recognition are displayed in \tab{tab:06task2}, with £500 awarded to joint 1\textsuperscript{st} (1\textsuperscript{st} \& 2\textsuperscript{nd}).

There is a good performance, with the best models (SDS-HD and SANO) both achieving 42\% on the task metric. The next top 2-models are not far behind, achieving $>34\%$ with the remaining bottom 2-models also not far behind, achieving $>27\%$.

The top two models use the well-known architecture of \gls{CNN} + \gls{LSTM} (+ Ensemble for SDS-HD). They are able to outperform purely spatial models (SK and GMAI) as well as more sophisticated models that utilise temporal decoders; positional encoding; and multi-task training (CITI and UNI-ANDES-23).

\begin{table}[!t]
\centering
\resizebox{0.60\columnwidth}{!}{
\begin{tabular}{c|c|c}
& Team & Macro-\F{} \\ \hline
1 & SDS-HD & 41.7±15.4 \\
2 & SANO & 41.6±06.3 \\
3 & CITI & 35.1±18.5 \\
4 & SK & 34.0±17.0 \\
5 & GMAI & 27.8±08.7 \\
6 & UNI-ANDES-23 & 27.5±13.5
\end{tabular}
}
\caption{19-instruments multi-label online recognition (task-2) rankings. Metrics are calculated across the 8-testing-videos (mean±std).}
\label{tab:06task2}
\end{table}

\begin{figure}[t!]
    \centering
    \includegraphics[width=\columnwidth]{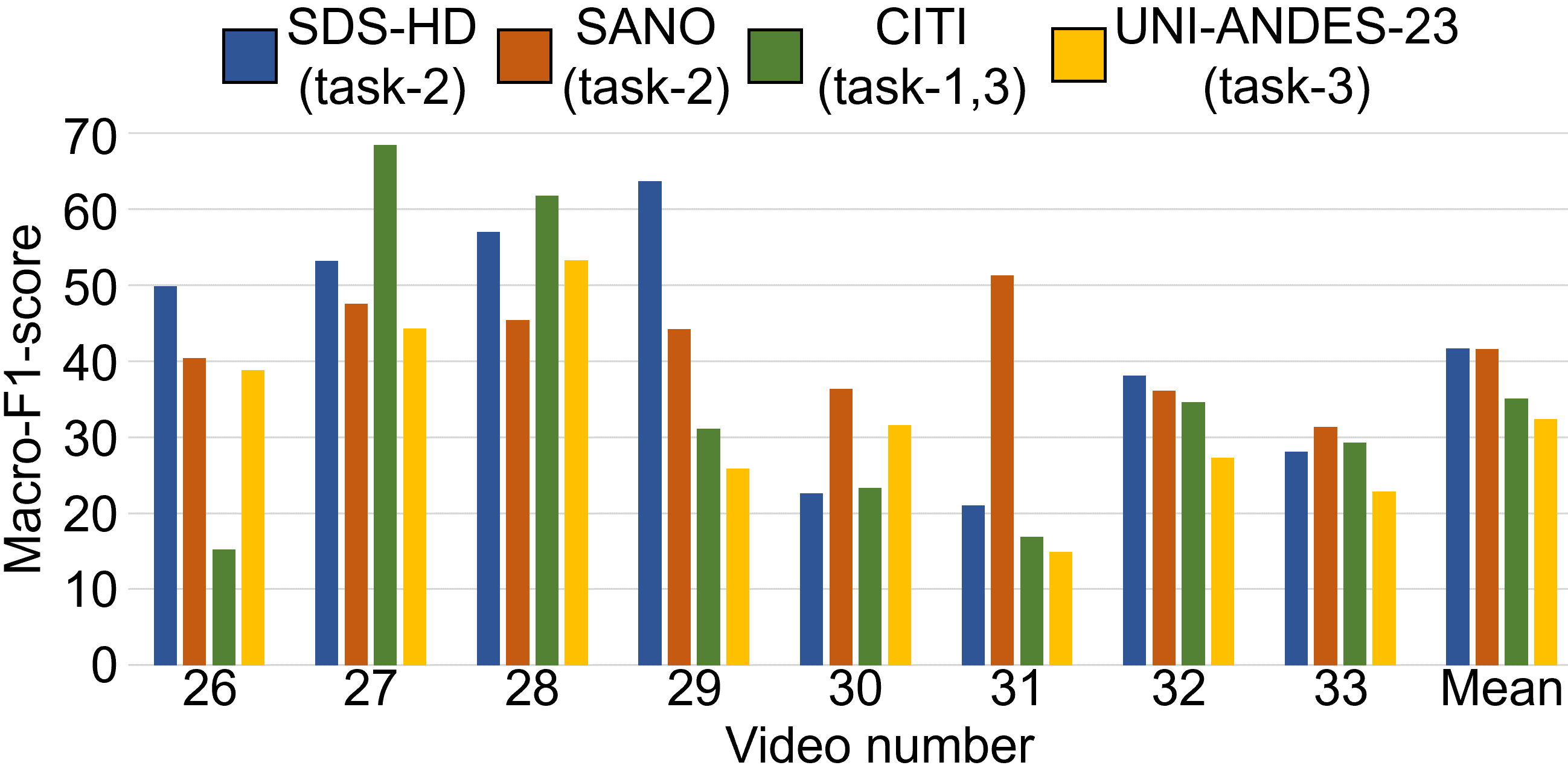}
    \caption{SDS-HD's (1\textsuperscript{st}) \& SANO's (2\textsuperscript{nd}) results for instrument recognition across the 8-testing-videos.}
    \label{fig:12instruments_test}
\end{figure}

\begin{figure*}[!t]
    \centering
    \begin{subfigure}[b]{0.49\textwidth}
        \centering
        \includegraphics[width=\columnwidth]{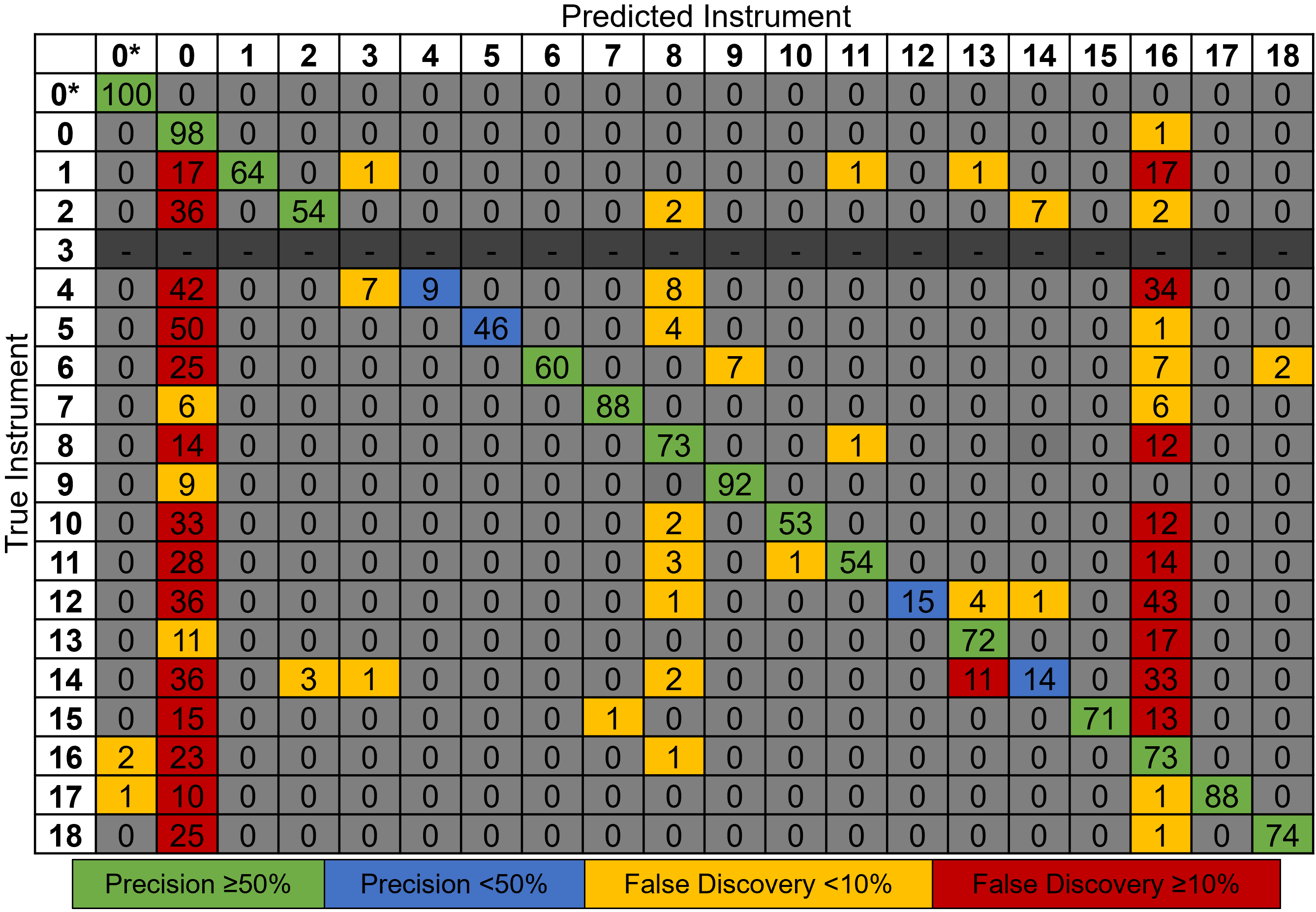}
        \caption{SDS-HD's task-2 (1\textsuperscript{st}) model.}
        \label{fig:13ainstruments_confusion_sds}
    \end{subfigure}
    \hfill
    \begin{subfigure}[b]{0.49\textwidth}
        \centering
        \includegraphics[width=\columnwidth]{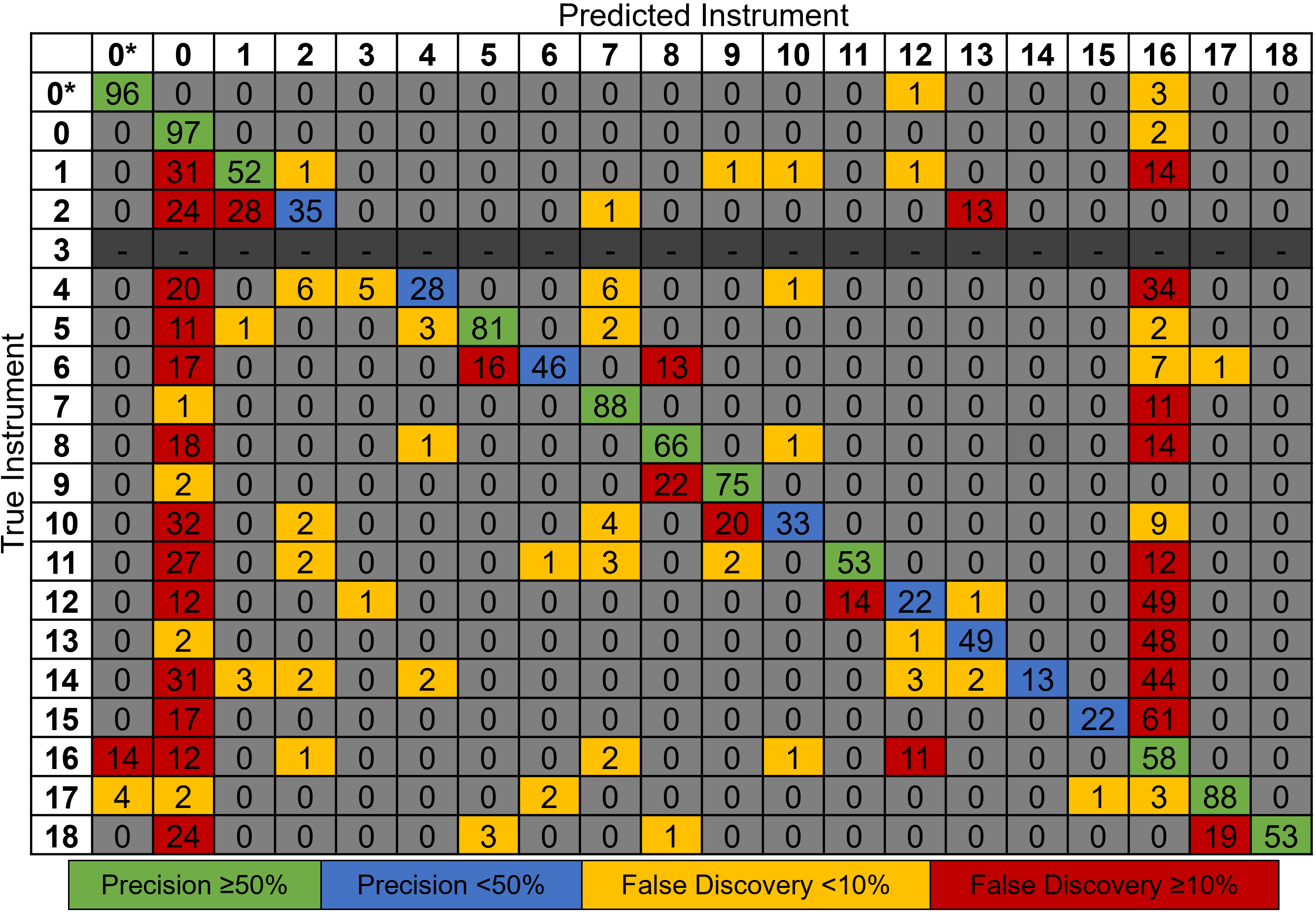}
        \caption{SANO's task-2 (2\textsuperscript{nd}) and task-3 (4\textsuperscript{th}) model.}
        \label{fig:13binstruments_confusion_sano}
    \end{subfigure}
        \begin{subfigure}[b]{0.49\textwidth}
        \centering
        \includegraphics[width=\columnwidth]{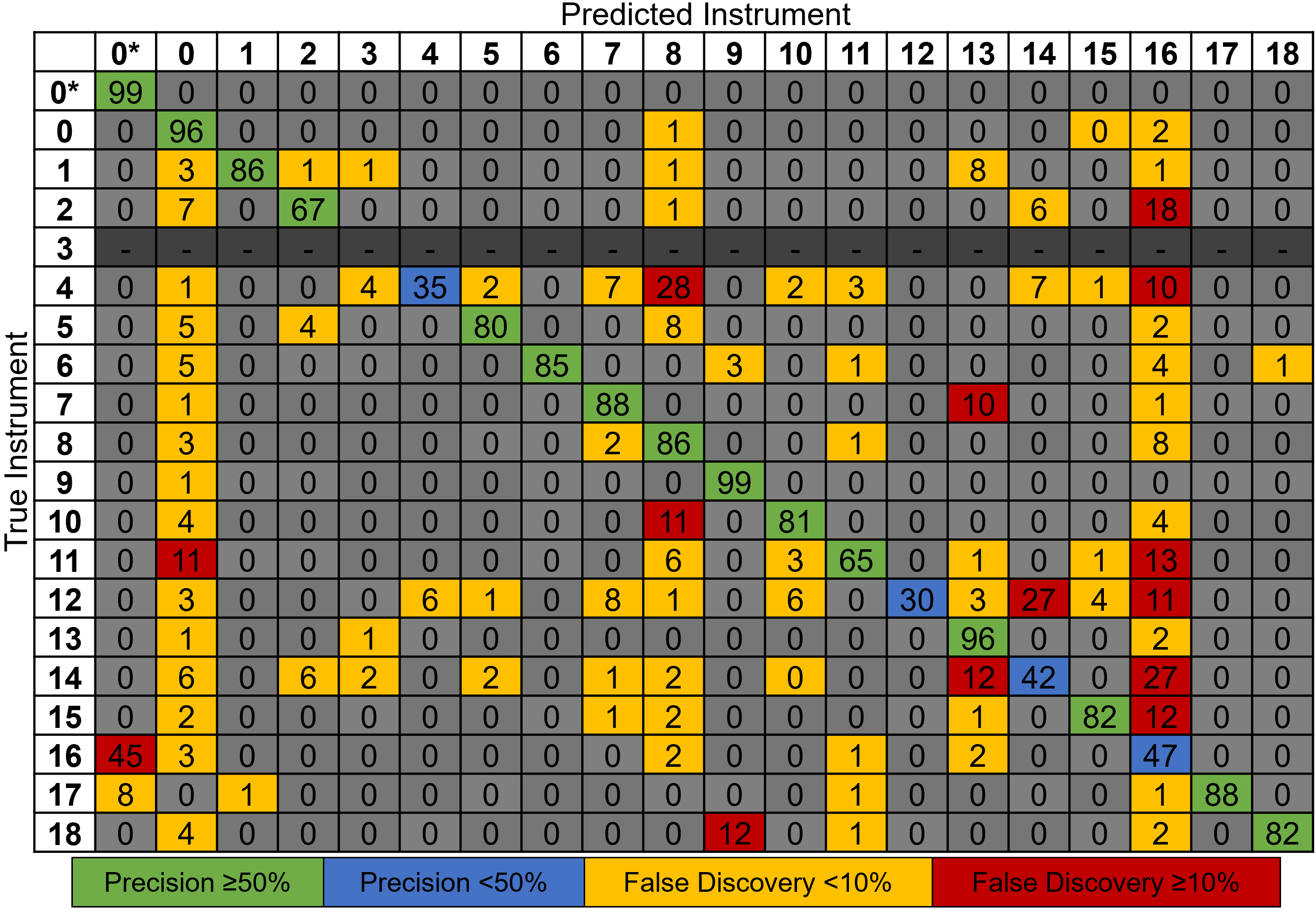}
        \caption{CITI's task-2 (3\textsuperscript{rd}) and task-3 (1\textsuperscript{st}) in model.}
        \label{fig:13cinstruments_confusion_citi}
    \end{subfigure}
    \hfill
    \begin{subfigure}[b]{0.49\textwidth}
        \centering
        \includegraphics[width=\columnwidth]{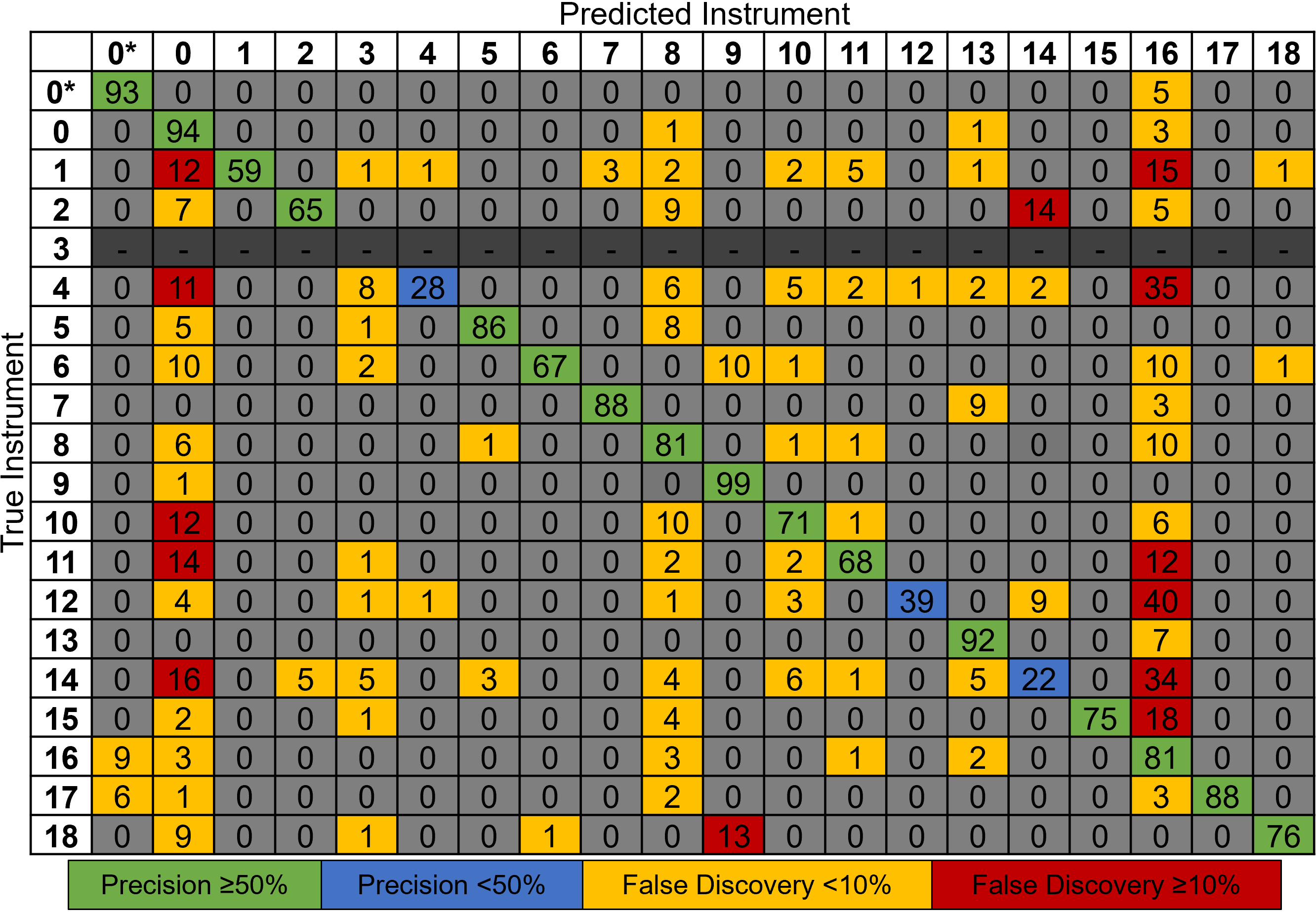}
        \caption{UNI-ANDES-23's task-3 (2\textsuperscript{nd}) model.}
        \label{fig:13dinstruments_confusion_uni}
    \end{subfigure}
    \caption{Instrument confusion matrices for the top models mean-averaged across the 8-testing-videos. 0* indicates \Q{no secondary instrument}. Instrument-3 (cup forceps) is not present in the testing dataset.}
    \label{fig:task2b}
\end{figure*}

There is varied \gls{std} in the top models as displayed in \fig{fig:12instruments_test}. SDS-HD outperforms the other models in the majority of videos. However, it is outperformed significantly by SANO in video-31 and by CITI in video-27. Like in step recognition, the video and model differences show the difficulty of a creating a generalised model.

\fig{fig:13ainstruments_confusion_sds} and \fig{fig:13binstruments_confusion_sano} displays the instrument confusion matrix for SDS-HD and SANO respectively. Instruments are frequently misclassified as instrument-0 (no instrument) and instrument-16 (suction). This is to be expected as they are the dominant classes, suggesting one way to overcome these incorrect predictions is through data balancing. Across both models, instruments 4; 12; 13 are predicted poorly with 2; 6; 10 also poorly predicted by SANO. This disparity is likely due to the number of instrument classes and the visual similarity between them, as well as insufficient training data. Interestingly, instruments 16 and 17, the only two secondary instruments in the testing dataset, are predicted well as secondary instruments. 

\subsection{Task-3}

\begin{table}[!t]
\centering
\resizebox{\columnwidth}{!}{
\begin{tabular}{c|c|c|c|c|c}
& \multirow{4}{*}{Team} & Step-(Macro-F\textsubscript{1} & \multirow{2}{*}{Step} & \multirow{2}{*}{Step} & \multirow{2}{*}{Instrument} \\
& & + Edit)/4 &  & &  \\
 &  & + Instrument- & Macro- & Edit- & Macro- \\ 
& & Macro-F\textsubscript{1}/2 & \F{} & score & \F{} \\ \hline
1 & CITI & 49.0±09.4 & 61.1±10.6 & 64.7±10.1 & 35.1±18.5 \\
\multirow{2}{*}{2} & UNI- & \multirow{2}{*}{40.5±07.7} & \multirow{2}{*}{51.0±08.8} & \multirow{2}{*}{46.3±10.4} & \multirow{2}{*}{32.4±11.7} \\
& ANDES-23 & & & & \\
3 & SK & 29.6±09.1 & 41.2±05.9 & 09.1±02.0 & 34.0±17.1 \\
4 & SANO & 28.3±06.4 & 39.6±06.5 & 01.4±00.4 & 36.2±14.8 \\
5 & GMAI & 15.5±03.6 & 07.2±00.7 & 00.5±00.1 & 27.2±06.9 \\
\end{tabular}
}
\caption{12-steps and 19-instruments multi-task online
recognition (task-3) rankings. Metrics are calculated across the 8-testing-videos (mean±std).}
\label{tab:07task3}
\end{table}

Results for the 5-submissions to  12-steps and 19-instruments multi-task online recognition are displayed in \tab{tab:07task3}, with £700 and £300 awarded to 1\textsuperscript{st} and 2\textsuperscript{nd} places respectively.

The performance is good, with the best models achieving 49\% (CITI) and 41\% (UNI-ANDES-23) on the task metric. The next top 2-models drop performance with $<30\%$, and the worst model only achieves 16\%. The \gls{std} is $<10\%$ across all models.

CITI's model is identical to its previous task models, which already utilised multi-task learning: the strong step recognition (1\textsuperscript{st}) compensates for the poorer instrument recognition (3\textsuperscript{rd}). On the other hand, UNI-ANDES-23's model improves in both step ($+0.4\%$) and instrument ($+4.9\%$) recognition due to the multi-task learning from the FusionTransformer. SK's instrument recognition model (4\textsuperscript{th}) now incorporates step recognition via an \gls{LSTM} achieving 25\% on task-1's metric, which would have given them 4\textsuperscript{th} place had they entered. SANO's model has decreased performance in both step ($-1.4\%$) and instrument ($-4.5\%$) recognition, this is due to their task-3 model not utilising the \gls{LSTM} trained for instrument recognition in task-2. GMAI's model performs similarly poorly in both step ($-0.2\%$) and instrument ($-0.6\%$) recognition. It is likely a multi-task form of TSO-NCT's model, which came 2\textsuperscript{nd} in task-1, would have performed well, given its similarity to the best models for instrument recognition. However, it is unlikely a multi-task form of DOLPHIN's and CAIR-POLYU-HK's task-1 models would have performed well given their poor performance in task-1.

The comparison of UNI-ANDES-23 task-3 model for each testing video is found in \fig{fig:11dsteps_test} (steps) and \fig{fig:12instruments_test} (instruments). For steps, is able to outperform TSO-NCT in videos 27; 30; 33, but is always outperformed by CITI. For instruments, it performs similarly to the other models, significantly outperforming CITI in video 26, although it is never the best performing model. 

\fig{fig:13cinstruments_confusion_citi} and \fig{fig:13dinstruments_confusion_uni} displays the instrument confusion matrix for CITI (1\textsuperscript{st}) and UNI-ANDES-23 (2\textsuperscript{nd}) respectively. When this is compared with the previously displayed confusion matrices, almost identical inferences can be made. One major difference is CITI overpredicts instrument-0 (no instrument) far less than other models, although it does overpredict instrument-0* (no secondary instrument) much more, reducing the precision of instrument-16 (suction). Similarly, \fig{fig:11csteps_confusion_uni} displays the the step confusion matrix for the UNI-ANDES-23. This is again similar to the previous matrices. Two minor differences are a poorer step-12 performance and a greater overprediction of step-14. 

\subsection{Benchmarks}
The 8-testing-videos are not released. Instead, top results of the suggested validation split are provided in \tab{tab:benchmark} to act as a benchmark for the community.

The best performing models on the suggested validation dataset for each metric are identical to the testing dataset, implying these models have good generalisation. This is more strongly true for step recognition, where there performance drop lower (-7\%) than instrument recognition (-47\%). This is likely due to overfitting to the small number of images of each minor instrument class.

\begin{table}[!t]
\centering
\resizebox{\columnwidth}{!}{
\begin{tabular}{c|c|c|c}
Team & Task-1 & Task-2 & Task-3 \\ \hline
CITI & \textbf{70} & 88 & \textbf{79} \\
SANO & 60 & 81 & 61 \\
SDS-HD & - & \textbf{89} & - \\
TSO-NCT & 67 & - & - \\
UNI-ANDES-23 & 69 & 79 & 71
\end{tabular}
}
\caption{Benchmark metric results for the suggested validation dataset, videos: 01, 12, 21, 24, 25. \textbf{Bold} indicates the best result for that column's task.}
\label{tab:benchmark}
\end{table}

\section{Conclusion} \label{sec:conclusion}
The \gls{PV}-2023 challenge pertains to developing deep learning models for workflow recognition for the \gls{eTSA}, with 3-tasks: (1) 12-step multi-class recognition; (2) 18-instrument multi-label recognition; and (3) 12-step and 18-instrument multi-task recognition. It was run across 5-months as a sub-challenge of the \gls{EV}-2023 challenge, with results and awards presented at the \gls{MICCAI}-2023 conference hosted in Vancouver, Canada on 08-Oct-2023. Participants were given access to the first curated public dataset of \gls{eTSA}: comprising 25-videos, with annotations for each second indicating the corresponding surgical step and instrument used. Across the 3-tasks there were 18-submissions from 9-teams across 6-countries.

The 9-models utilise a variety of state-of-the-art computer vision and workflow recognition techniques and architectures. Training techniques include random augmentations; end-to-end training; multi-task training; and data balancing. Architectures are generally split into 3-stages. Stage-1 consists of a encoder: either purely spatial via a \gls{CNN} or \gls{S-TF}; or spatial-temporal via a \gls{ST-TF}. Stage-2, if used, consists of a \gls{ST-D}: either a \gls{LSTM} or \gls{ST-TF}. Stage-3, if used, consists of a online post-processing technique, usually a \gls{TSF}. Some models also utilise ensembles. Performance was found to be strong for both established architectures (e.g. \gls{CNN} + \gls{LSTM} + \gls{TSF}) as well as less established custom architectures utilising temporal propogation. A commonality between the best architectures was the use of a \gls{ST-D} and \gls{TSF}. 

This challenge provides benchmark performances for workflow recognition in \gls{eTSA}, overcoming many of the difficulties previously outlined. Some of these difficulties, however, still need to be overcome before the predictions are reliable enough to be used in clinical practice. Other important factors to consider are: explainability of models, which is essential for a clinical setting; environmental impacts of model training, as some models were trained for long periods of time; and real-time implementation, which was enforced as models had to run at $10\times$ speed on the 32-GB GPU.

This challenge was limited primarily by the difficulty of data acquisition: obtaining consent; recording videos; and annotating videos. A larger multi-centered dataset would allow for improved generalisability of models. Although the challenge has ended, the website will remain, and the data is publicly available, along with the benchmark results. Future work will include: refining existing and trialing new models to address \gls{eTSA} specific difficulties; and transfer learning from foundational models trained on alternative publicly available minimally-invasive datasets.  

The Pituitary Vision 2023 Challenge showcases the efforts of the international minimally invasive surgical computer vision community on endoscopic pituitary surgery. The models created not only verify their generalisability on a new dataset, but advance the field, pushing it closer to usable clinical assistance.

\clearpage

\section*{Declarations}
\subsection*{Acknowledgements}
The authors would like to thank the \gls{EV}-2023 organisation committee for running the grand challenge and the \gls{MICCAI}-2023 committee for hosting the conference. With thanks to Digital Surgery Ltd, a Medtronic company, for access to Touch Surgery Ecosystem for video recording, annotation, and storage.

\subsection*{Funding}
The \gls{PV} challenge was funded by Digital Surgery, Medtronic. This work was supported in whole, or in part, by the \gls{WEISS} [203145/Z/16/Z], the \gls{EPSRC} [EP/W00805X/1, EP/Y01958X/1, EP/P012841/1], the Horizon 2020 FET [GA863146], the Department of Science, Innovation and Technology (DSIT) and the Royal Academy of Engineering under the Chair in Emerging Technologies programme. Adrito Das is supported by the \gls{EPSRC} [EP/S021612/1]. Danyal Z. Khan is supported by a \gls{NIHR} Academic Clinical Fellowship and the \gls{CRUK} Pre-doctoral Fellowship. John G. Hanrahan is supported by a \gls{NIHR} Academic Clinical Fellowship. Hani J. Marcus is supported by \gls{WEISS} [NS/A000050/1] and by the \gls{NIHR} Biomedical Research Centre at \gls{UCL}. 

\subsection*{Contributions}
Adrito Das, Danyal Z. Khan, Dimitrios Psychogyios, Yitong Zhang, John G. Hanrahan, Francisco Vasconcelos, Sophia Bano, Hani J. Marcus, and Danail Stoyanov organised the PitVis challenge. Adrito Das was the primary organiser of the challenge. Danyal Z. Khan, John G. Hanrahan, and Hani J. Marcus facilitated the recording and annotating of the endoscopic pituitary videos. Dimitrios Psychogyios created and maintained the challenge website. Yitong Zhang created the baseline models. Francisco Vasconcelos and Danail Stoyanov provided the resources to run the models. Sophia Bano provided the supervision throughout the challenge organisation.

Adrito Das wrote the original draft of this paper. The rest of the organisation team reviewed and edited the paper. All other authors were participants in the challenge and reviewed their respective team sections. A maximum of three authors per participating team was permitted.

\subsection*{Ethics} 
The study was registered with \gls{UCL} \gls{IRB} (17819/011). 

\subsection*{Data and publishing}
The data for this challenge cannot be distributed but is available under a CC BY-NC-SA 4.0 license: \url{www.doi.org/10.5522/04/26531686}. Data used in the challenge can be used for publication purposes only after the joint publication summarising the challenge results is published. For the purpose of open access, the author has applied a CC-BY public copyright licence to any author accepted manuscript version arising from this submission.

\subsection*{Conflict of interest} 
The authors declare that they have no conflict of interest.

\clearpage

\onecolumn{
\bibliography{main}

\begin{thebibliography}{47}
\providecommand{\natexlab}[1]{#1}
\providecommand{\url}[1]{\texttt{#1}}
\expandafter\ifx\csname urlstyle\endcsname\relax
  \providecommand{\doi}[1]{doi: #1}\else
  \providecommand{\doi}{doi: \begingroup \urlstyle{rm}\Url}\fi

\bibitem[Maier-Hein et~al.(2022)Maier-Hein, Eisenmann, Sarikaya, M\"{a}rz, Collins, Malpani, Fallert, Feussner, Giannarou, Mascagni, Nakawala, Park, Pugh, Stoyanov, Vedula, Cleary, Fichtinger, Forestier, Gibaud, Grantcharov, Hashizume, Heckmann-N\"{o}tzel, Kenngott, Kikinis, M\"{u}ndermann, Navab, Onogur, Roß, Sznitman, Taylor, Tizabi, Wagner, Hager, Neumuth, Padoy, Collins, Gockel, Goedeke, Hashimoto, Joyeux, Lam, Leff, Madani, Marcus, Meireles, Seitel, Teber, \"{U}ckert, M\"{u}ller-Stich, Jannin, and Speidel]{MaierHein2022}
Lena Maier-Hein, Matthias Eisenmann, Duygu Sarikaya, Keno M\"{a}rz, Toby Collins, Anand Malpani, Johannes Fallert, Hubertus Feussner, Stamatia Giannarou, Pietro Mascagni, Hirenkumar Nakawala, Adrian Park, Carla Pugh, Danail Stoyanov, Swaroop~S. Vedula, Kevin Cleary, Gabor Fichtinger, Germain Forestier, Bernard Gibaud, Teodor Grantcharov, Makoto Hashizume, Doreen Heckmann-N\"{o}tzel, Hannes~G. Kenngott, Ron Kikinis, Lars M\"{u}ndermann, Nassir Navab, Sinan Onogur, Tobias Roß, Raphael Sznitman, Russell~H. Taylor, Minu~D. Tizabi, Martin Wagner, Gregory~D. Hager, Thomas Neumuth, Nicolas Padoy, Justin Collins, Ines Gockel, Jan Goedeke, Daniel~A. Hashimoto, Luc Joyeux, Kyle Lam, Daniel~R. Leff, Amin Madani, Hani~J. Marcus, Ozanan Meireles, Alexander Seitel, Dogu Teber, Frank \"{U}ckert, Beat~P. M\"{u}ller-Stich, Pierre Jannin, and Stefanie Speidel.
\newblock Surgical data science – from concepts toward clinical translation.
\newblock \emph{Medical Image Analysis}, 76:\penalty0 102306, February 2022.
\newblock ISSN 1361-8415.
\newblock \doi{10.1016/j.media.2021.102306}.
\newblock URL \url{http://dx.doi.org/10.1016/j.media.2021.102306}.

\bibitem[Speidel et~al.(2023)Speidel, Maier-Hein, Stoyanov, Bodenstedt, Reinke, Bano, Jenke, Wagner, Daum, Tabibian, {Adrito Das}, {Yitong Zhang}, Vasconcelos, Psychogyios, {Danyal Z. Khan}, Marcus, {Aneeq Zia}, Liu, Bhattacharyya, {Ziheng Wang}, Berniker, Perreault, Jarc, Malpani, Glock, {Haozheng Xu}, Xu, {Baoru Huang}, and Giannarou]{EndoVis2023}
Stefanie Speidel, Lena Maier-Hein, Danail Stoyanov, Sebastian Bodenstedt, Annika Reinke, Sophia Bano, Alexander Jenke, Martin Wagner, Marie Daum, Ala Tabibian, {Adrito Das}, {Yitong Zhang}, Francisco Vasconcelos, Dimitris Psychogyios, {Danyal Z. Khan}, Hani~J. Marcus, {Aneeq Zia}, Xi~Liu, Kiran Bhattacharyya, {Ziheng Wang}, Max Berniker, Conor Perreault, Anthony Jarc, Anand Malpani, Kimberly Glock, {Haozheng Xu}, Chi Xu, {Baoru Huang}, and Stamatia Giannarou.
\newblock Endoscopic vision challenge 2023, 2023.
\newblock URL \url{https://zenodo.org/record/8315050}.

\bibitem[Ganapathy and Tadi(2022)]{Ganapathy2022}
Muthu~Kuzhali Ganapathy and Prasanna Tadi.
\newblock Anatomy, head and neck, pituitary gland.
\newblock \emph{StatPearls [Internet]}, July 2022.
\newblock \doi{http://www.ncbi.nlm.nih.gov/books/NBK551529/}.
\newblock \url{http://www.ncbi.nlm.nih.gov/books/NBK551529/} (accessed Aug 2024).

\bibitem[Russ et~al.(2022)Russ, Anastasopoulou, and Shafiq]{Russ2022}
Sophia Russ, Catherine Anastasopoulou, and Ismat Shafiq.
\newblock Pituitary adenoma.
\newblock \emph{StatPearls [Internet]}, July 2022.
\newblock \doi{https://www.ncbi.nlm.nih.gov/books/NBK554451/}.
\newblock \url{https://www.ncbi.nlm.nih.gov/books/NBK554451/} (accessed Aug 2024).

\bibitem[Agustsson et~al.(2015)Agustsson, Baldvinsdottir, Jonasson, Olafsdottir, Steinthorsdottir, Sigurdsson, Thorsson, Carroll, Korbonits, and Benediktsson]{Agustsson2015}
Tomas~Thor Agustsson, Tinna Baldvinsdottir, Jon~G Jonasson, Elinborg Olafsdottir, Valgerdur Steinthorsdottir, Gunnar Sigurdsson, Arni~V Thorsson, Paul~V Carroll, Márta Korbonits, and Rafn Benediktsson.
\newblock The epidemiology of pituitary adenomas in iceland, 1955–2012: a nationwide population-based study.
\newblock \emph{European Journal of Endocrinology}, 173\penalty0 (5):\penalty0 655–664, November 2015.
\newblock ISSN 1479-683X.
\newblock \doi{10.1530/eje-15-0189}.
\newblock URL \url{http://dx.doi.org/10.1530/eje-15-0189}.

\bibitem[Ogra et~al.(2014)Ogra, Nichols, Stylli, Kaye, Savino, and Danesh-Meyer]{Ogra2014}
Siddharth Ogra, Andrew~D. Nichols, Stanley Stylli, Andrew~H. Kaye, Peter~J. Savino, and Helen~V. Danesh-Meyer.
\newblock Visual acuity and pattern of visual field loss at presentation in pituitary adenoma.
\newblock \emph{Journal of Clinical Neuroscience}, 21\penalty0 (5):\penalty0 735–740, May 2014.
\newblock ISSN 0967-5868.
\newblock \doi{10.1016/j.jocn.2014.01.005}.
\newblock URL \url{http://dx.doi.org/10.1016/j.jocn.2014.01.005}.

\bibitem[Tritos and Biller(2019)]{Tritos2019}
Nicholas~A. Tritos and Beverly~M.K. Biller.
\newblock Medical management of cushing disease.
\newblock \emph{Neurosurgery Clinics of North America}, 30\penalty0 (4):\penalty0 499–508, October 2019.
\newblock ISSN 1042-3680.
\newblock \doi{10.1016/j.nec.2019.05.007}.
\newblock URL \url{http://dx.doi.org/10.1016/j.nec.2019.05.007}.

\bibitem[Wang et~al.(2014)Wang, Zhou, Wei, Meng, Zhang, Hou, and Sun]{Wang2014}
Fuyu Wang, Tao Zhou, Shaobo Wei, Xianghui Meng, Jiashu Zhang, Yuanzheng Hou, and Guochen Sun.
\newblock Endoscopic endonasal transsphenoidal surgery of 1, 166 pituitary adenomas.
\newblock \emph{Surgical Endoscopy}, 29\penalty0 (6):\penalty0 1270–1280, October 2014.
\newblock ISSN 1432-2218.
\newblock \doi{10.1007/s00464-014-3815-0}.
\newblock URL \url{http://dx.doi.org/10.1007/s00464-014-3815-0}.

\bibitem[Marcus et~al.(2021)Marcus, Khan, Borg, Buchfelder, Cetas, Collins, Dorward, Fleseriu, Gurnell, Javadpour, Jones, Koh, Layard~Horsfall, Mamelak, Mortini, Muirhead, Oyesiku, Schwartz, Sinha, Stoyanov, Syro, Tsermoulas, Williams, Winder, Zada, and Laws]{Marcus2021}
Hani~J. Marcus, Danyal~Z. Khan, Anouk Borg, Michael Buchfelder, Justin~S. Cetas, Justin~W. Collins, Neil~L. Dorward, Maria Fleseriu, Mark Gurnell, Mohsen Javadpour, Pamela~S. Jones, Chan~Hee Koh, Hugo Layard~Horsfall, Adam~N. Mamelak, Pietro Mortini, William Muirhead, Nelson~M. Oyesiku, Theodore~H. Schwartz, Saurabh Sinha, Danail Stoyanov, Luis~V. Syro, Georgios Tsermoulas, Adam Williams, Mark~J. Winder, Gabriel Zada, and Edward~R. Laws.
\newblock Pituitary society expert delphi consensus: operative workflow in endoscopic transsphenoidal pituitary adenoma resection.
\newblock \emph{Pituitary}, 24\penalty0 (6):\penalty0 839–853, July 2021.
\newblock ISSN 1573-7403.
\newblock \doi{10.1007/s11102-021-01162-3}.
\newblock URL \url{http://dx.doi.org/10.1007/s11102-021-01162-3}.

\bibitem[Consortium(2023)]{Cranial2023}
CRANIAL Consortium.
\newblock Machine learning driven prediction of cerebrospinal fluid rhinorrhoea following endonasal skull base surgery: A multicentre prospective observational study.
\newblock \emph{Frontiers in Oncology}, 13, March 2023.
\newblock ISSN 2234-943X.
\newblock \doi{10.3389/fonc.2023.1046519}.
\newblock URL \url{http://dx.doi.org/10.3389/fonc.2023.1046519}.

\bibitem[Frara et~al.(2020)Frara, Rodriguez-Carnero, Formenti, Martinez-Olmos, Giustina, and Casanueva]{Frara2020}
Stefano Frara, Gemma Rodriguez-Carnero, Ana~M. Formenti, Miguel~A. Martinez-Olmos, Andrea Giustina, and Felipe~F. Casanueva.
\newblock Pituitary tumors centers of excellence.
\newblock \emph{Endocrinology and Metabolism Clinics of North America}, 49\penalty0 (3):\penalty0 553–564, September 2020.
\newblock ISSN 0889-8529.
\newblock \doi{10.1016/j.ecl.2020.05.010}.
\newblock URL \url{http://dx.doi.org/10.1016/j.ecl.2020.05.010}.

\bibitem[Wang et~al.(2022)Wang, Sun, Liu, and Gu]{Wang2022}
Yan Wang, Qiyuan Sun, Zhenzhong Liu, and Lin Gu.
\newblock Visual detection and tracking algorithms for minimally invasive surgical instruments: A comprehensive review of the state-of-the-art.
\newblock \emph{Robotics and Autonomous Systems}, 149:\penalty0 103945, March 2022.
\newblock ISSN 0921-8890.
\newblock \doi{10.1016/j.robot.2021.103945}.
\newblock URL \url{http://dx.doi.org/10.1016/j.robot.2021.103945}.

\bibitem[Khan et~al.(2022)Khan, Luengo, Barbarisi, Addis, Culshaw, Dorward, Haikka, Jain, Kerr, Koh, Layard~Horsfall, Muirhead, Palmisciano, Vasey, Stoyanov, and Marcus]{Khan2022}
Danyal~Z. Khan, Imanol Luengo, Santiago Barbarisi, Carole Addis, Lucy Culshaw, Neil~L. Dorward, Pinja Haikka, Abhiney Jain, Karen Kerr, Chan~Hee Koh, Hugo Layard~Horsfall, William Muirhead, Paolo Palmisciano, Baptiste Vasey, Danail Stoyanov, and Hani~J. Marcus.
\newblock Automated operative workflow analysis of endoscopic pituitary surgery using machine learning: development and preclinical evaluation (ideal stage 0).
\newblock \emph{Journal of Neurosurgery}, 137\penalty0 (1):\penalty0 51–58, July 2022.
\newblock ISSN 1933-0693.
\newblock \doi{10.3171/2021.6.jns21923}.
\newblock URL \url{http://dx.doi.org/10.3171/2021.6.jns21923}.

\bibitem[Khan et~al.(2023)Khan, Hanrahan, Baldeweg, Dorward, Stoyanov, and Marcus]{Khan2023}
Danyal~Z Khan, John~G Hanrahan, Stephanie~E Baldeweg, Neil~L Dorward, Danail Stoyanov, and Hani~J Marcus.
\newblock Current and future advances in surgical therapy for pituitary adenoma.
\newblock \emph{Endocrine Reviews}, 44\penalty0 (5):\penalty0 947–959, May 2023.
\newblock ISSN 1945-7189.
\newblock \doi{10.1210/endrev/bnad014}.
\newblock URL \url{http://dx.doi.org/10.1210/endrev/bnad014}.

\bibitem[Khan et~al.(2024{\natexlab{a}})Khan, Newall, Koh, Das, Aapan, Horsfall, Baldeweg, Bano, Borg, Chari, Dorward, Elserius, Giannis, Jain, Stoyanov, and Marcus]{Khan2024B}
Danyal~Z. Khan, Nicola Newall, Chan~Hee Koh, Adrito Das, Sanchit Aapan, Hugo~Layard Horsfall, Stephanie~E. Baldeweg, Sophia Bano, Anouk Borg, Aswin Chari, Neil~L. Dorward, Anne Elserius, Theofanis Giannis, Abhiney Jain, Danail Stoyanov, and Hani~J. Marcus.
\newblock Video-based performance analysis in pituitary surgery - part 2: Artificial intelligence assisted surgical coaching.
\newblock \emph{World Neurosurgery}, August 2024{\natexlab{a}}.
\newblock ISSN 1878-8750.
\newblock \doi{10.1016/j.wneu.2024.07.219}.
\newblock URL \url{http://dx.doi.org/10.1016/j.wneu.2024.07.219}.

\bibitem[Das et~al.(2023{\natexlab{a}})Das, Khan, Hanrahan, Marcus, and Stoyanov]{Das2023}
Adrito Das, Danyal~Z. Khan, John~G. Hanrahan, Hani~J. Marcus, and Danail Stoyanov.
\newblock Automatic generation of operation notes in endoscopic pituitary surgery videos using workflow recognition.
\newblock \emph{Intelligence-Based Medicine}, 8:\penalty0 100107, 2023{\natexlab{a}}.
\newblock ISSN 2666-5212.
\newblock \doi{10.1016/j.ibmed.2023.100107}.
\newblock URL \url{http://dx.doi.org/10.1016/j.ibmed.2023.100107}.

\bibitem[He et~al.(2024)He, Xu, Das, Khan, Bano, Marcus, Stoyanov, Clarkson, and Islam]{He2024}
Runlong He, Mengya Xu, Adrito Das, Danyal~Z. Khan, Sophia Bano, Hani~J. Marcus, Danail Stoyanov, Matthew~J. Clarkson, and Mobarakol Islam.
\newblock Pitvqa: Image-grounded text embedding llm for visual question answering in pituitary surgery, 2024.
\newblock URL \url{https://arxiv.org/abs/2405.13949}.

\bibitem[Khan et~al.(2024{\natexlab{b}})Khan, Koh, Das, Valetopolou, Hanrahan, Horsfall, Baldeweg, Bano, Borg, Dorward, Olukoya, Stoyanov, and Marcus]{Khan2024A}
Danyal~Z. Khan, Chan~Hee Koh, Adrito Das, Alexandra Valetopolou, John~G. Hanrahan, Hugo~Layard Horsfall, Stephanie~E. Baldeweg, Sophia Bano, Anouk Borg, Neil~L. Dorward, Olatomiwa Olukoya, Danail Stoyanov, and Hani~J. Marcus.
\newblock Video-based performance analysis in pituitary surgery - part 1: Surgical outcomes.
\newblock \emph{World Neurosurgery}, August 2024{\natexlab{b}}.
\newblock ISSN 1878-8750.
\newblock \doi{10.1016/j.wneu.2024.07.218}.
\newblock URL \url{http://dx.doi.org/10.1016/j.wneu.2024.07.218}.

\bibitem[Garrow et~al.(2020)Garrow, Kowalewski, Li, Wagner, Schmidt, Engelhardt, Hashimoto, Kenngott, Bodenstedt, Speidel, M\"{u}ller-Stich, and Nickel]{Garrow2020}
Carly~R. Garrow, Karl-Friedrich Kowalewski, Linhong Li, Martin Wagner, Mona~W. Schmidt, Sandy Engelhardt, Daniel~A. Hashimoto, Hannes~G. Kenngott, Sebastian Bodenstedt, Stefanie Speidel, Beat~P. M\"{u}ller-Stich, and Felix Nickel.
\newblock Machine learning for surgical phase recognition: A systematic review.
\newblock \emph{Annals of Surgery}, 273\penalty0 (4):\penalty0 684–693, November 2020.
\newblock ISSN 1528-1140.
\newblock \doi{10.1097/sla.0000000000004425}.
\newblock URL \url{http://dx.doi.org/10.1097/sla.0000000000004425}.

\bibitem[Maier-Hein et~al.(2020)Maier-Hein, Reinke, Kozubek, Martel, Arbel, Eisenmann, Hanbury, Jannin, M\"{u}ller, Onogur, Saez-Rodriguez, van Ginneken, Kopp-Schneider, and Landman]{MaierHein2020}
Lena Maier-Hein, Annika Reinke, Michal Kozubek, Anne~L. Martel, Tal Arbel, Matthias Eisenmann, Allan Hanbury, Pierre Jannin, Henning M\"{u}ller, Sinan Onogur, Julio Saez-Rodriguez, Bram van Ginneken, Annette Kopp-Schneider, and Bennett~A. Landman.
\newblock Bias: Transparent reporting of biomedical image analysis challenges.
\newblock \emph{Medical Image Analysis}, 66:\penalty0 101796, December 2020.
\newblock ISSN 1361-8415.
\newblock \doi{10.1016/j.media.2020.101796}.
\newblock URL \url{http://dx.doi.org/10.1016/j.media.2020.101796}.

\bibitem[Rueckert et~al.(2024)Rueckert, Rueckert, and Palm]{Rueckert2024}
Tobias Rueckert, Daniel Rueckert, and Christoph Palm.
\newblock Methods and datasets for segmentation of minimally invasive surgical instruments in endoscopic images and videos: A review of the state of the art.
\newblock \emph{Computers in Biology and Medicine}, 169:\penalty0 107929, February 2024.
\newblock ISSN 0010-4825.
\newblock \doi{10.1016/j.compbiomed.2024.107929}.
\newblock URL \url{http://dx.doi.org/10.1016/j.compbiomed.2024.107929}.

\bibitem[Demir et~al.(2023)Demir, Schieber, Weise, Roth, May, Maier, and Yang]{Demir2023}
Kubilay~Can Demir, Hannah Schieber, Tobias Weise, Daniel Roth, Matthias May, Andreas Maier, and Seung~Hee Yang.
\newblock Deep learning in surgical workflow analysis: A review of phase and step recognition.
\newblock \emph{IEEE Journal of Biomedical and Health Informatics}, 27\penalty0 (11):\penalty0 5405–5417, November 2023.
\newblock ISSN 2168-2208.
\newblock \doi{10.1109/jbhi.2023.3311628}.
\newblock URL \url{http://dx.doi.org/10.1109/JBHI.2023.3311628}.

\bibitem[Maier-Hein et~al.(2024)Maier-Hein, Reinke, Godau, Tizabi, Buettner, Christodoulou, Glocker, Isensee, Kleesiek, Kozubek, Reyes, Riegler, Wiesenfarth, Kavur, Sudre, Baumgartner, Eisenmann, Heckmann-N\"{o}tzel, R\"{a}dsch, Acion, Antonelli, Arbel, Bakas, Benis, Blaschko, Cardoso, Cheplygina, Cimini, Collins, Farahani, Ferrer, Galdran, van Ginneken, Haase, Hashimoto, Hoffman, Huisman, Jannin, Kahn, Kainmueller, Kainz, Karargyris, Karthikesalingam, Kofler, Kopp-Schneider, Kreshuk, Kurc, Landman, Litjens, Madani, Maier-Hein, Martel, Mattson, Meijering, Menze, Moons, M\"{u}ller, Nichyporuk, Nickel, Petersen, Rajpoot, Rieke, Saez-Rodriguez, Sánchez, Shetty, van Smeden, Summers, Taha, Tiulpin, Tsaftaris, Van~Calster, Varoquaux, and J\"{a}ger]{MaierHein2024}
Lena Maier-Hein, Annika Reinke, Patrick Godau, Minu~D. Tizabi, Florian Buettner, Evangelia Christodoulou, Ben Glocker, Fabian Isensee, Jens Kleesiek, Michal Kozubek, Mauricio Reyes, Michael~A. Riegler, Manuel Wiesenfarth, A.~Emre Kavur, Carole~H. Sudre, Michael Baumgartner, Matthias Eisenmann, Doreen Heckmann-N\"{o}tzel, Tim R\"{a}dsch, Laura Acion, Michela Antonelli, Tal Arbel, Spyridon Bakas, Arriel Benis, Matthew~B. Blaschko, M.~Jorge Cardoso, Veronika Cheplygina, Beth~A. Cimini, Gary~S. Collins, Keyvan Farahani, Luciana Ferrer, Adrian Galdran, Bram van Ginneken, Robert Haase, Daniel~A. Hashimoto, Michael~M. Hoffman, Merel Huisman, Pierre Jannin, Charles~E. Kahn, Dagmar Kainmueller, Bernhard Kainz, Alexandros Karargyris, Alan Karthikesalingam, Florian Kofler, Annette Kopp-Schneider, Anna Kreshuk, Tahsin Kurc, Bennett~A. Landman, Geert Litjens, Amin Madani, Klaus Maier-Hein, Anne~L. Martel, Peter Mattson, Erik Meijering, Bjoern Menze, Karel G.~M. Moons, Henning M\"{u}ller, Brennan Nichyporuk, Felix Nickel,
  Jens Petersen, Nasir Rajpoot, Nicola Rieke, Julio Saez-Rodriguez, Clara~I. Sánchez, Shravya Shetty, Maarten van Smeden, Ronald~M. Summers, Abdel~A. Taha, Aleksei Tiulpin, Sotirios~A. Tsaftaris, Ben Van~Calster, Gaël Varoquaux, and Paul~F. J\"{a}ger.
\newblock Metrics reloaded: recommendations for image analysis validation.
\newblock \emph{Nature Methods}, 21\penalty0 (2):\penalty0 195–212, February 2024.
\newblock ISSN 1548-7105.
\newblock \doi{10.1038/s41592-023-02151-z}.
\newblock URL \url{http://dx.doi.org/10.1038/s41592-023-02151-z}.

\bibitem[Das et~al.(2022)Das, Bano, Vasconcelos, Khan, Marcus, and Stoyanov]{Das2022}
Adrito Das, Sophia Bano, Francisco Vasconcelos, Danyal~Z. Khan, Hani~J Marcus, and Danail Stoyanov.
\newblock Reducing prediction volatility in the surgical workflow recognition of endoscopic pituitary surgery.
\newblock \emph{International Journal of Computer Assisted Radiology and Surgery}, 17\penalty0 (8):\penalty0 1445–1452, April 2022.
\newblock ISSN 1861-6429.
\newblock \doi{10.1007/s11548-022-02599-y}.
\newblock URL \url{http://dx.doi.org/10.1007/s11548-022-02599-y}.

\bibitem[Twinanda et~al.(2017)Twinanda, Shehata, Mutter, Marescaux, de~Mathelin, and Padoy]{Twinanda2017}
Andru~P. Twinanda, Sherif Shehata, Didier Mutter, Jacques Marescaux, Michel de~Mathelin, and Nicolas Padoy.
\newblock Endonet: A deep architecture for recognition tasks on laparoscopic videos.
\newblock \emph{IEEE Transactions on Medical Imaging}, 36\penalty0 (1):\penalty0 86–97, January 2017.
\newblock ISSN 1558-254X.
\newblock \doi{10.1109/tmi.2016.2593957}.
\newblock URL \url{http://dx.doi.org/10.1109/TMI.2016.2593957}.

\bibitem[Psychogyios et~al.(2024)Psychogyios, Colleoni, Van~Amsterdam, Li, Huang, Li, Jia, Zou, Wang, Liu, Boels, Huo, Sparks, Dasgupta, Granados, Ourselin, Xu, Wang, Wu, Bai, Ren, Yamada, Harai, Ishikawa, Hayashi, Simoens, DeBacker, Cisternino, Furnari, Mottrie, Ferraguti, Kondo, Kasai, Hirasawa, Kim, Lee, Lee, Kong, Fu, Li, An, Krell, Bodenstedt, Ayobi, Perez, Rodriguez, Puentes, Arbelaez, Mohareri, and Stoyanov]{Psychogyios2024}
Dimitrios Psychogyios, Emanuele Colleoni, Beatrice Van~Amsterdam, Chih-Yang Li, Shu-Yu Huang, Yuchong Li, Fucang Jia, Baosheng Zou, Guotai Wang, Yang Liu, Maxence Boels, Jiayu Huo, Rachel Sparks, Prokar Dasgupta, Alejandro Granados, Sebastien Ourselin, Mengya Xu, An~Wang, Yanan Wu, Long Bai, Hongliang Ren, Atsushi Yamada, Yuriko Harai, Yuto Ishikawa, Kazuyuki Hayashi, Jente Simoens, Pieter DeBacker, Francesco Cisternino, Gabriele Furnari, Alex Mottrie, Federica Ferraguti, Satoshi Kondo, Satoshi Kasai, Kousuke Hirasawa, Soohee Kim, Seung~Hyun Lee, Kyu~Eun Lee, Hyoun-Joong Kong, Kui Fu, Chao Li, Shan An, Stefanie Krell, Sebastian Bodenstedt, Nicolas Ayobi, Alejandra Perez, Santiago Rodriguez, Juanita Puentes, Pablo Arbelaez, Omid Mohareri, and Danail Stoyanov.
\newblock Sar-rarp50: Segmentation of surgical instrumentation and action recognition on robot-assisted radical prostatectomy challenge, 2024.
\newblock URL \url{https://arxiv.org/abs/2401.00496}.

\bibitem[Alabi et~al.(2024)Alabi, Vercauteren, and Shi]{Oluwatosin2024}
Oluwatosin Alabi, Tom Vercauteren, and Miaojing Shi.
\newblock Multitask learning in minimally invasive surgical vision: A review, 2024.
\newblock URL \url{https://arxiv.org/abs/2401.08256}.

\bibitem[Das et~al.(2023{\natexlab{b}})Das, Khan, Williams, Hanrahan, Borg, Dorward, Bano, Marcus, and Stoyanov]{Das2023B}
Adrito Das, Danyal~Z. Khan, Simon~C. Williams, John~G. Hanrahan, Anouk Borg, Neil~L. Dorward, Sophia Bano, Hani~J. Marcus, and Danail Stoyanov.
\newblock \emph{A Multi-task Network for Anatomy Identification in Endoscopic Pituitary Surgery}, page 472–482.
\newblock Springer Nature Switzerland, 2023{\natexlab{b}}.
\newblock ISBN 9783031439964.
\newblock \doi{10.1007/978-3-031-43996-4_45}.
\newblock URL \url{http://dx.doi.org/10.1007/978-3-031-43996-4_45}.

\bibitem[Mao et~al.(2024)Mao, Das, Islam, Khan, Williams, Hanrahan, Borg, Dorward, Clarkson, Stoyanov, Marcus, and Bano]{Mao2024}
Zhehua Mao, Adrito Das, Mobarakol Islam, Danyal~Z. Khan, Simon~C. Williams, John~G. Hanrahan, Anouk Borg, Neil~L. Dorward, Matthew~J. Clarkson, Danail Stoyanov, Hani~J. Marcus, and Sophia Bano.
\newblock Pitsurgrt: real-time localization of critical anatomical structures in endoscopic pituitary surgery.
\newblock \emph{International Journal of Computer Assisted Radiology and Surgery}, 19\penalty0 (6):\penalty0 1053–1060, March 2024.
\newblock ISSN 1861-6429.
\newblock \doi{10.1007/s11548-024-03094-2}.
\newblock URL \url{http://dx.doi.org/10.1007/s11548-024-03094-2}.

\bibitem[Jin et~al.(2020)Jin, Li, Dou, Chen, Qin, Fu, and Heng]{Jin2020}
Yueming Jin, Huaxia Li, Qi~Dou, Hao Chen, Jing Qin, Chi-Wing Fu, and Pheng-Ann Heng.
\newblock Multi-task recurrent convolutional network with correlation loss for surgical video analysis.
\newblock \emph{Medical Image Analysis}, 59:\penalty0 101572, January 2020.
\newblock ISSN 1361-8415.
\newblock \doi{10.1016/j.media.2019.101572}.
\newblock URL \url{http://dx.doi.org/10.1016/j.media.2019.101572}.

\bibitem[Lea et~al.(2016)Lea, Reiter, Vidal, and Hager]{Lea2016}
Colin Lea, Austin Reiter, René Vidal, and Gregory~D. Hager.
\newblock \emph{Segmental Spatiotemporal CNNs for Fine-Grained Action Segmentation}, page 36–52.
\newblock Springer International Publishing, 2016.
\newblock ISBN 9783319464879.
\newblock \doi{10.1007/978-3-319-46487-9_3}.
\newblock URL \url{http://dx.doi.org/10.1007/978-3-319-46487-9_3}.

\bibitem[Zou et~al.(2022)Zou, Liu, Wang, Tao, and Zheng]{Zou2022}
Xiaoyang Zou, Wenyong Liu, Junchen Wang, Rong Tao, and Guoyan Zheng.
\newblock Arst: auto-regressive surgical transformer for phase recognition from laparoscopic videos.
\newblock \emph{Computer Methods in Biomechanics and Biomedical Engineering: Imaging \&; Visualization}, 11\penalty0 (4):\penalty0 1012–1018, November 2022.
\newblock ISSN 2168-1171.
\newblock \doi{10.1080/21681163.2022.2145238}.
\newblock URL \url{http://dx.doi.org/10.1080/21681163.2022.2145238}.

\bibitem[Bochkovskiy et~al.(2020)Bochkovskiy, Wang, and Liao]{Bochkovskiy2020}
Alexey Bochkovskiy, Chien-Yao Wang, and Hong-Yuan~Mark Liao.
\newblock Yolov4: Optimal speed and accuracy of object detection, 2020.
\newblock URL \url{https://arxiv.org/abs/2004.10934}.

\bibitem[Liu et~al.(2022)Liu, Mao, Wu, Feichtenhofer, Darrell, and Xie]{Liu2022}
Zhuang Liu, Hanzi Mao, Chao-Yuan Wu, Christoph Feichtenhofer, Trevor Darrell, and Saining Xie.
\newblock A convnet for the 2020s, 2022.
\newblock URL \url{https://arxiv.org/abs/2201.03545}.

\bibitem[Huang et~al.(2016)Huang, Liu, van~der Maaten, and Weinberger]{Huang2016}
Gao Huang, Zhuang Liu, Laurens van~der Maaten, and Kilian~Q. Weinberger.
\newblock Densely connected convolutional networks, 2016.
\newblock URL \url{https://arxiv.org/abs/1608.06993}.

\bibitem[Zhang et~al.(2022)Zhang, Li, Liu, Zhang, Su, Zhu, Ni, and Shum]{Zhang2022}
Hao Zhang, Feng Li, Shilong Liu, Lei Zhang, Hang Su, Jun Zhu, Lionel~M. Ni, and Heung-Yeung Shum.
\newblock Dino: Detr with improved denoising anchor boxes for end-to-end object detection, 2022.
\newblock URL \url{https://arxiv.org/abs/2203.03605}.

\bibitem[Tan and Le(2019)]{Tan2020}
Mingxing Tan and Quoc~V. Le.
\newblock Efficientnet: Rethinking model scaling for convolutional neural networks, 2019.
\newblock URL \url{https://arxiv.org/abs/1905.11946}.

\bibitem[Fang et~al.(2024)Fang, Sun, Wang, Huang, Wang, and Cao]{Fang2024}
Yuxin Fang, Quan Sun, Xinggang Wang, Tiejun Huang, Xinlong Wang, and Yue Cao.
\newblock Eva-02: A visual representation for neon genesis.
\newblock \emph{Image and Vision Computing}, 149:\penalty0 105171, September 2024.
\newblock ISSN 0262-8856.
\newblock \doi{10.1016/j.imavis.2024.105171}.
\newblock URL \url{http://dx.doi.org/10.1016/j.imavis.2024.105171}.

\bibitem[Fan et~al.(2021)Fan, Xiong, Mangalam, Li, Yan, Malik, and Feichtenhofer]{Fan2021}
Haoqi Fan, Bo~Xiong, Karttikeya Mangalam, Yanghao Li, Zhicheng Yan, Jitendra Malik, and Christoph Feichtenhofer.
\newblock Multiscale vision transformers.
\newblock In \emph{2021 IEEE/CVF International Conference on Computer Vision (ICCV)}. IEEE, October 2021.
\newblock \doi{10.1109/iccv48922.2021.00675}.
\newblock URL \url{http://dx.doi.org/10.1109/ICCV48922.2021.00675}.

\bibitem[He et~al.(2016)He, Zhang, Ren, and Sun]{He2016}
Kaiming He, Xiangyu Zhang, Shaoqing Ren, and Jian Sun.
\newblock Deep residual learning for image recognition.
\newblock In \emph{2016 IEEE Conference on Computer Vision and Pattern Recognition (CVPR)}. IEEE, June 2016.
\newblock \doi{10.1109/cvpr.2016.90}.
\newblock URL \url{http://dx.doi.org/10.1109/CVPR.2016.90}.

\bibitem[Liu et~al.(2021)Liu, Lin, Cao, Hu, Wei, Zhang, Lin, and Guo]{Liu2021}
Ze~Liu, Yutong Lin, Yue Cao, Han Hu, Yixuan Wei, Zheng Zhang, Stephen Lin, and Baining Guo.
\newblock Swin transformer: Hierarchical vision transformer using shifted windows.
\newblock In \emph{2021 IEEE/CVF International Conference on Computer Vision (ICCV)}. IEEE, October 2021.
\newblock \doi{10.1109/iccv48922.2021.00986}.
\newblock URL \url{http://dx.doi.org/10.1109/ICCV48922.2021.00986}.

\bibitem[Czempiel et~al.(2020)Czempiel, Paschali, Keicher, Simson, Feussner, Kim, and Navab]{Czempiel2020}
Tobias Czempiel, Magdalini Paschali, Matthias Keicher, Walter Simson, Hubertus Feussner, Seong~Tae Kim, and Nassir Navab.
\newblock \emph{TeCNO: Surgical Phase Recognition with Multi-stage Temporal Convolutional Networks}, page 343–352.
\newblock Springer International Publishing, 2020.
\newblock ISBN 9783030597160.
\newblock \doi{10.1007/978-3-030-59716-0_33}.
\newblock URL \url{http://dx.doi.org/10.1007/978-3-030-59716-0_33}.

\bibitem[Wu et~al.(2022{\natexlab{a}})Wu, Zhang, Peng, Liu, Xiao, Fu, and Yuan]{Wu2022}
Kan Wu, Jinnian Zhang, Houwen Peng, Mengchen Liu, Bin Xiao, Jianlong Fu, and Lu~Yuan.
\newblock Tinyvit: Fast pretraining distillation for small vision transformers, 2022{\natexlab{a}}.
\newblock URL \url{https://arxiv.org/abs/2207.10666}.

\bibitem[El-Nouby et~al.(2021)El-Nouby, Touvron, Caron, Bojanowski, Douze, Joulin, Laptev, Neverova, Synnaeve, Verbeek, and Jegou]{Elnouby2021}
Alaaeldin El-Nouby, Hugo Touvron, Mathilde Caron, Piotr Bojanowski, Matthijs Douze, Armand Joulin, Ivan Laptev, Natalia Neverova, Gabriel Synnaeve, Jakob Verbeek, and Hervé Jegou.
\newblock Xcit: Cross-covariance image transformers, 2021.
\newblock URL \url{https://arxiv.org/abs/2106.09681}.

\bibitem[Zou et~al.(2024)Zou, Yu, Tao, and Zheng]{Zou2024}
Xiaoyang Zou, Derong Yu, Rong Tao, and Guoyan Zheng.
\newblock \emph{An End-to-End Spatial-Temporal Transformer Model for Surgical Action Triplet Recognition}, page 114–120.
\newblock Springer Nature Switzerland, 2024.
\newblock ISBN 9783031514852.
\newblock \doi{10.1007/978-3-031-51485-2_14}.
\newblock URL \url{http://dx.doi.org/10.1007/978-3-031-51485-2_14}.

\bibitem[Wu et~al.(2022{\natexlab{b}})Wu, Zhang, Peng, Liu, Xiao, Fu, and Yuan]{Vaswani2017}
Kan Wu, Jinnian Zhang, Houwen Peng, Mengchen Liu, Bin Xiao, Jianlong Fu, and Lu~Yuan.
\newblock Tinyvit: Fast pretraining distillation for small vision transformers, 2022{\natexlab{b}}.
\newblock URL \url{https://arxiv.org/abs/2207.10666}.

\bibitem[Ban et~al.(2021)Ban, Rosman, Ward, Hashimoto, Kondo, Iwaki, Meireles, and Rus]{Ban2021}
Yutong Ban, Guy Rosman, Thomas Ward, Daniel Hashimoto, Taisei Kondo, Hidekazu Iwaki, Ozanan Meireles, and Daniela Rus.
\newblock Aggregating long-term context for learning laparoscopic and robot-assisted surgical workflows.
\newblock In \emph{2021 IEEE International Conference on Robotics and Automation (ICRA)}. IEEE, May 2021.
\newblock \doi{10.1109/icra48506.2021.9561770}.
\newblock URL \url{http://dx.doi.org/10.1109/ICRA48506.2021.9561770}.

\end{thebibliography}
}
\onecolumn{
\setlist[description]{itemsep=0pt, topsep=0pt, parsep=0pt, partopsep=0pt}
\glsnogroupskiptrue
\printglossary[type=\acronymtype]
}
\end{document}